\documentclass{article}
\usepackage{iclr2026_conference,times}

\usepackage{amsmath,amsfonts,bm}

\def\eqref#1{equation~\ref{#1}}

\def\1{\bm{1}}

\DeclareMathAlphabet{\mathsfit}{\encodingdefault}{\sfdefault}{m}{sl}
\SetMathAlphabet{\mathsfit}{bold}{\encodingdefault}{\sfdefault}{bx}{n}

\usepackage{amsmath}
\usepackage{amssymb}
\usepackage{graphicx}
\usepackage{booktabs}
\usepackage{multirow}
\usepackage{nicefrac}
\usepackage{wrapfig}
\usepackage{subcaption}
\usepackage{placeins}
\usepackage{tikz}
\usepackage{xcolor}
\usepackage{hyperref}
\usepackage{url}
\usepackage{cleveref}

\title{The Hyperspherical Geometry of CLIP Latent Space: A Semantic Mixture Model}

\author{\small Zijie Yu$^1$, Gaowen Liu$^2$, Ramana Rao Kompella$^2$, Philip S. Yu$^3$, Yue Song$^1$\thanks{Denotes corresponding author.} \\
\small $^1$Tsinghua University, $^2$Cisco Research, $^3$University of Illinois Chicago}

\iclrfinalcopy 
\begin{document}

\maketitle
\lhead{}
\renewcommand{\headrulewidth}{0pt}

\begin{abstract}
Contrastive Language--Image Pretraining (CLIP) representations form a semantic embedding space governed by cosine similarity, reflecting an intrinsic hyperspherical geometry. However, existing probabilistic interpretations typically rely on Gaussian assumptions, which fail to capture this directional and multimodal structure. We propose a principled density model for the CLIP latent space based on Mixtures of von Mises--Fisher (MovMF) distributions defined on the unit hypersphere. Using the Expectation--Maximization (EM) algorithm, we efficiently learn a probabilistic model in which each mixture component corresponds to a coherent semantic concept. This formulation yields a closed-form likelihood naturally aligned with hyperspherical geometry, enabling accurate and interpretable density estimation. Empirically, our model significantly improves long-tailed and out-of-distribution detection and provides a natural semantic decomposition, representing each embedding as a sparse probabilistic combination of interpretable concepts. These results suggest that CLIP latent space is more faithfully characterized as a hyperspherical semantic mixture rather than an isotropic Gaussian, establishing a simple and geometrically consistent probabilistic framework for modeling and understanding multimodal representations. Project page is available at \url{https://xiaoyuzhizi.github.io/movmf-clip/}.
\end{abstract}

\section{Introduction}
\label{sec:intro}






Contrastive Language–Image Pre-training (CLIP)~\citep{radford2021learningtransferablevisualmodels} has become the foundation of modern vision–language models (VLMs). By aligning visual and textual representations in a shared embedding space, CLIP enables a broad range of downstream tasks, including zero-shot recognition~\citep{radford2021learningtransferablevisualmodels} and text-to-image generation~\citep{ramesh2022hierarchicaltextconditionalimagegeneration}. Since semantic similarity is measured by cosine distance, the latent space is often idealized as a normalized hypersphere with isotropic geometry.

Despite this theoretical simplicity, the empirical geometry of CLIP representations deviates substantially from the isotropic ideal. Prior studies reveal pronounced structural irregularities, including the \emph{Modality Gap}~\citep{liang2022mindgapunderstandingmodality} and the so-called \emph{Cone Effect}~\citep{liang2022mindgapunderstandingmodality}, where embeddings concentrate within restricted angular regions. Such anisotropy distorts angular similarity, biases representations toward dominant concepts such as large objects or spurious background features, and degrades robustness across long-tail categories~\citep{abbasi2025clipmicroscopefinegrainedanalysis,wang2024soberlookrobustnessclips,shao2024examining}.

\begin{figure}[t]
    \centering
    \begin{subfigure}[b]{0.45\textwidth}
        \centering
        \includegraphics[width=\textwidth]{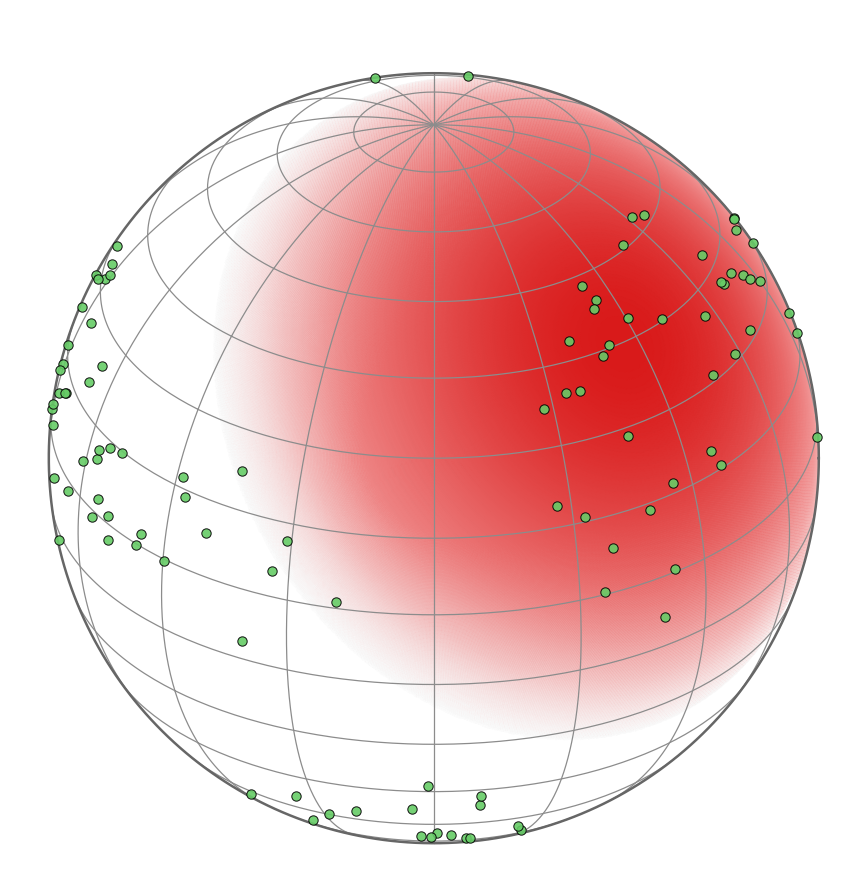}
        \caption{Global Gaussian in $\mathbb{R}^d$}
        \label{fig:teaser_wclip}
    \end{subfigure}
    \hfill
    \begin{subfigure}[b]{0.45\textwidth}
        \centering
        \includegraphics[width=\textwidth]{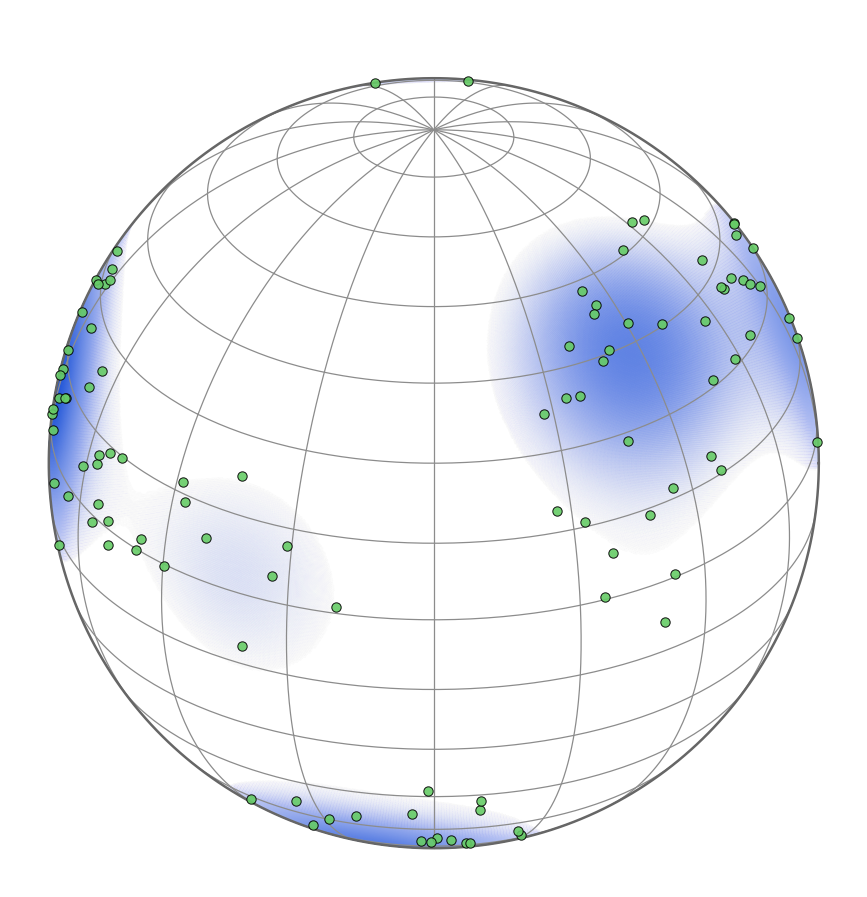}
        \caption{MovMF-CLIP on $\mathbb{S}^{d-1}$}
        \label{fig:teaser_movmf}
    \end{subfigure}
    \caption{\textbf{Density modeling of CLIP latent space (visualized via dimensionality reduction on real data).} \textbf{(a)} Gaussian-based approaches such as W-CLIP model the latent space with a single global distribution, which can assign low likelihood to valid long-tail concepts due to their distance from the global mean. \textbf{(b)} MovMF-CLIP models the space as a hyperspherical semantic mixture, capturing the multimodal structures and providing density estimates aligned with semantic clusters.}
    \label{fig:teaser}
\end{figure}

To enable probabilistic reasoning in this space, recent work has proposed transforming CLIP embeddings into a Gaussianized coordinate system. In particular, Whitened CLIP (W-CLIP)~\citep{betser2025whitenedcliplikelihoodsurrogate} applies an invertible whitening transformation and models the resulting features with a single isotropic Gaussian, yielding a tractable likelihood based on Euclidean magnitude. While effective, this unimodal assumption is fundamentally misaligned with the directional and multimodal nature of contrastive embeddings. Natural semantic concepts form multiple structured regions on the hypersphere. Modeling the space with a single Gaussian inevitably conflates \emph{semantic rarity} with \emph{distributional abnormality}, assigning low likelihood to rare but valid concepts and treating them similarly to true out-of-distribution (OOD) samples. Moreover, a global scalar likelihood provides limited interpretability of the underlying semantic structure. 
It can only indicate how typical a sample is with respect to the entire distribution, but does not reveal which semantic modes contribute to that score or how the representation decomposes across distinct concepts.

In this work, we propose \textbf{MovMF-CLIP}, a geometrically consistent probabilistic framework that models CLIP latent space as a \emph{hyperspherical semantic mixture}. We first perform covariance-based geometric calibration via whitening to remove global anisotropy while preserving semantic relationships. Then we model the normalized representations using a Mixture of von Mises–Fisher (MovMF) distributions, estimated efficiently via the Expectation–Maximization (EM) algorithm. Unlike Gaussian approximations that rely on magnitude, MovMF components are defined intrinsically on the unit hypersphere and capture the multimodal directional structures (see Fig.~\ref{fig:teaser}). Each mixture component corresponds to a coherent semantic prototype, yielding a closed-form likelihood that respects the geometry of contrastive embeddings. This probabilistic formulation supports two complementary capabilities. First, it provides a principled density estimate for likelihood-based evaluation, decoupling semantic rarity from low probability and improving robustness in long-tailed and OOD settings. Second, the mixture structure enables intrinsic interpretability: embeddings can be decomposed into sparse probabilistic combinations of semantic prototypes. 

Empirically, we validate MovMF-CLIP across long-tailed fairness analysis, OOD detection, semantic decomposition, and semantic stability under iterative generative drift. 
For OOD detection, MovMF-CLIP reduces FPR95 from 67.76\% to \textbf{48.00\%} on MS-COCO, and from 75.05\% to \textbf{33.48\%} solely on tail concepts. For semantic decomposition, MovMF-CLIP achieves the highest Semantic Relevance of \textbf{0.673} while achieving a \textbf{13 $\times$ speedup} in inference time compared to the second-best method. Across tasks, modeling the latent space as a hyperspherical semantic mixture consistently improves robustness, interpretability, and stability. 

Our main contributions are summarized as follows:

\begin{itemize}
    \item We offer a geometric re-interpretation of CLIP latent space, arguing that unimodal Gaussian modeling is inherently inconsistent with its directional hypersphere geometry and multimodal semantic structure.
    \item We propose \textbf{MovMF-CLIP}, an elegant and principled density modeling framework that integrates covariance-based geometric calibration with von Mises–Fisher mixture estimation on the unit sphere.
    \item We show that this unified geometric–probabilistic formulation simultaneously enables (i) calibrated likelihood estimation robust to long-tail and OOD settings, (ii) intrinsic and efficient probabilistic semantic decomposition without auxiliary decoders, and (iii) a lightweight geometric prior for stabilizing multimodal representations under iterative generative drift.
\end{itemize}

\section{Related Work}
\subsection{Geometry and Anisotropy in CLIP}
CLIP and related multimodal contrastive models~\citep{radford2021learningtransferablevisualmodels,li2022blipbootstrappinglanguageimagepretraining,yu2022cocacontrastivecaptionersimagetext} embed images and text into a shared latent space optimized under cosine similarity. This objective implicitly normalizes embeddings onto a hypersphere and is often interpreted as encouraging an isotropic and uniformly distributed representation geometry~\citep{wang2022understandingcontrastiverepresentationlearning}. However, a growing body of empirical evidence demonstrates that the actual geometry of CLIP latent space deviates substantially from this idealized picture.

One prominent deviation is the \emph{Modality Gap}, wherein image and text embeddings occupy distinct regions of the hypersphere rather than forming a fully overlapping distribution~\citep{liang2022mindgapunderstandingmodality,levi2025doubleellipsoidgeometryclip}. While initially viewed as a deficiency in cross-modal alignment, subsequent studies suggest that such separation may serve functional roles, including improved robustness and mitigation of catastrophic forgetting~\citep{huang2025mindgappreservingcompensating}. Beyond modality-level separation, CLIP representations also exhibit significant anisotropy and uneven semantic coverage. Embeddings tend to concentrate around dominant visual patterns, such as large foreground objects or frequently co-occurring features, while underrepresenting rare or fine-grained concepts~\citep{abbasi2025clipmicroscopefinegrainedanalysis,wang2024soberlookrobustnessclips,lan2024clearclipdecomposingcliprepresentations}. The anisotropic structure leads to substantial performance variability across semantics, particularly on long-tailed and rare concepts~\citep{tu2024closerlookrobustnesscontrastive,shao2024examining}. 

These observations indicate that CLIP latent space possesses structured and non-uniform density rather than a single homogeneous distribution. Consequently, cosine similarity alone cannot fully capture semantic relationships, as it ignores variations in representation density and concentration~\citep{Steck_2024,kang2025clipidealnofix}. Several works have therefore explored alternative geometric perspectives, including cycle-consistency constraints~\citep{goel2022cyclipcycliccontrastivelanguageimage} and optimal transport formulations~\citep{shi2024otclip}. More recent analyses explicitly characterize CLIP geometry as structured anisotropy, such as the ``double-ellipsoid'' structure separating common and rare concepts~\citep{levi2025doubleellipsoidgeometryclip,wen2024makescliprobustlongtailed}. Collectively, these findings suggest that CLIP latent space exhibits an inherently multimodal semantic organization.

To enable likelihood-based reasoning and improve calibration, recent work has proposed geometric normalization techniques. W-CLIP~\citep{betser2025whitenedcliplikelihoodsurrogate}, for example, applies a whitening transformation to normalize second-order statistics and estimate likelihood under a Gaussian assumption. Related approaches similarly attempt to regularize representation geometry through linear normalization~\citep{chung2026globalgeometryvisionrepresentations}. While effective for calibration, such methods rely on unimodal approximations of the latent space. In contrast, our work directly models CLIP latent space as a mixture of hyperspherical semantic components, providing a probabilistic formulation that better captures its intrinsic multimodal structure.

\subsection{Likelihood Estimation in Latent Space}

Estimating likelihood directly in pixel space is computationally challenging and often poorly aligned with semantic similarity, motivating likelihood estimation in learned representation spaces. Recent work has explored using pretrained multimodal encoders such as CLIP to define semantically meaningful likelihood surrogates. In particular, W-CLIP~\citep{betser2025whitenedcliplikelihoodsurrogate} projects embeddings into a whitened space and estimates likelihood under a Gaussian assumption, providing a tractable likelihood proxy. However, modeling the latent space with a single Gaussian imposes a unimodal assumption that conflicts with the inherently multimodal structure of semantic representations. Natural image distributions consist of multiple semantic modes corresponding to different objects, attributes, and compositions. Under a unimodal Gaussian model, rare but valid concepts may be assigned artificially low likelihood, conflating semantic rarity with distributional abnormality. Alternative approaches, such as retrieval-based similarity~\citep{He_2025_ICCV}, prompt adaptation~\citep{zhou2026anomalyclipobjectagnosticpromptlearning,Cao_2024}, or architectural modifications~\citep{gong2024feclip}, can improve robustness but do not provide a unified probabilistic formulation.

In contrast, we model the CLIP latent distribution using a Mixture of von Mises–Fisher distributions, which naturally respects the hyperspherical geometry of CLIP embeddings. This formulation captures multiple semantic modes while retaining analytical tractability for likelihood estimation.

\subsection{Interpretability of Multimodal Representations}

Understanding and interpreting dense representations learned by multimodal models has become an important area of research. Prior work has explored sparse linear decompositions that express embeddings as combinations of interpretable concepts drawn from predefined vocabularies or learned dictionaries~\citep{bhalla2024interpretingclipsparselinear,hoang-xuan2025advancing,parekh2024conceptbasedexplainabilityframeworklarge}. Other approaches investigate internal model structure, including attention mechanisms and sparse autoencoders, to identify latent features and disentangle semantic factors within CLIP representations~\citep{gandelsman2024interpretingclipsimagerepresentation,kempf2025doesclipenabledomain,zaigrajew2025interpretingcliphierarchicalsparse,dhimoila2026crossmodalredundancygeometryvisionlanguage}. While these methods provide valuable insights, they often require training auxiliary models, introducing additional complexity and dependence on external supervision or architectural modifications.

In contrast, our approach derives interpretability directly from probabilistic modeling of the latent space. By representing embeddings as a mixture of hyperspherical distributions, each component corresponds to a coherent semantic mode, and individual embeddings can be interpreted through their probabilistic associations with these components. The mixture centers naturally act as semantic prototypes, enabling semantic explanations and likelihood estimation within a unified framework, without auxiliary decoders or external concept dictionaries.
\section{MovMF-CLIP: Hyperspherical Density Modeling of CLIP Latent Space}

\begin{figure}[t]
    \centering
    \includegraphics[width=0.99\linewidth]{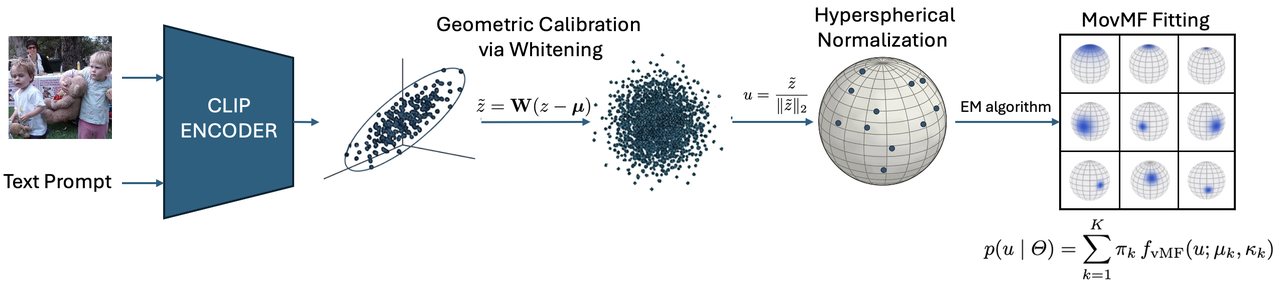} 
    \caption{\textbf{Overview of the MovMF-CLIP Framework.} We first extract raw embeddings using the CLIP encoder, which exhibit severe anisotropy. To address this, we apply geometric calibration via whitening ($\tilde{z} = \mathbf{W}(z - \boldsymbol{\mu})$) and normalize the features onto a unit hypersphere ($u = \tilde{z}/\|\tilde{z}\|_2$). Finally, we fit a MovMF distributions on the hypersphere using the EM algorithm, yielding a principled multimodal density model.}
    \label{fig:pipeline}
    \vspace{-3mm}
\end{figure}

We model CLIP latent space as a geometrically calibrated hyperspherical density framework. As shown in Fig.~\ref{fig:pipeline}, our framework proceeds in three stages: 
(i) metric normalization via whitening to remove global anisotropy (Sec.~\ref{sec:prelim}); (ii) multimodal directional density modeling using MovMF with EM estimation (Secs.~\ref{sec:movmf_model}–\ref{sec:em_algo}); and 
(iii) probabilistic inference enabling likelihood-based evaluation and semantic attribution (Sec.~\ref{sec:inference}). 
Together, these components form a unified geometric–probabilistic framework for modeling CLIP representations.

\subsection{Geometric Calibration of CLIP Latent Space}
\label{sec:prelim}

We consider a dataset $\mathcal{D} = \{(x_i^{\text{img}}, x_i^{\text{text}})\}_{i=1}^N$ consisting of image–text pairs. A pre-trained CLIP model provides two encoders, $f_{\text{img}}(\cdot)$ and $f_{\text{text}}(\cdot)$, which map each modality into a shared $d$-dimensional embedding space. We denote by $z \in \mathbb{R}^d$ the latent representation produced by either encoder. Our geometric calibration and density modeling are applied independently to each modality within this shared space. Although CLIP is trained with cosine similarity and is interpreted as operating on a hypersphere, the raw latent distribution is empirically far from geometrically isotropic. In particular, the embeddings exhibit strong second-order anisotropy and dominant covariance directions, reflecting dataset bias and representation artifacts. As a result, angular distances in the raw space are distorted by global variance structure, potentially obscuring the true semantic organization of the data.

\noindent\textbf{Geometric Calibration via Whitening.} To remove global anisotropy while preserving semantic structure, we first apply a whitening transformation in $\mathbb{R}^d$. Let $\boldsymbol{\mu}$ denote the empirical mean and $\boldsymbol{\Sigma}$ the covariance matrix computed over the reference dataset (\emph{e.g.} $5k$ validation samples of MS-COCO). We construct a whitening operator $\mathbf{W} = \boldsymbol{\Sigma}^{-1/2}$ and define the calibrated embedding
\begin{equation}
    \tilde{z} = \mathbf{W}(z - \boldsymbol{\mu}).
\end{equation}
This transformation standardizes second-order statistics so that $\mathrm{Cov}(\tilde{z}) \approx \mathbf{I}$. Unlike likelihood-based approaches such as W-CLIP~\citep{betser2025whitenedcliplikelihoodsurrogate}, we do not assume that the whitened representations follow a Gaussian distribution. Instead, whitening serves purely as a metric normalization step: it removes nuisance covariance structure and \textbf{induces an intrinsic Mahalanobis geometry} in the original latent space (see Appendix B). Consequently, directional comparisons are no longer dominated by high-variance axes unrelated to semantic content.

\noindent\textbf{Hyperspherical Geometry.} CLIP representations are inherently interpreted through cosine similarity, meaning that their semantics are encoded directionally on the unit hypersphere. Accordingly, after geometric calibration, we operate on normalized embeddings
\begin{equation}
    u = \frac{\tilde{z}}{\|\tilde{z}\|_2}, \qquad u \in \mathbb{S}^{d-1},
\end{equation}
which respects the intrinsic hyperspherical geometry imposed by contrastive training. This step does not introduce an additional modeling assumption; rather, it aligns our probabilistic framework with the geometry already underlying CLIP. By separating global covariance normalization in $\mathbb{R}^d$ from directional modeling on $\mathbb{S}^{d-1}$, we obtain a geometrically consistent representation that isolates semantic structure from nuisance anisotropy. This calibrated hyperspherical representation forms the foundation for our mixture-based density modeling.

\subsection{Modeling Latent Density via MovMF}
\label{sec:movmf_model}

Following the geometric calibration defined in Sec.~\ref{sec:prelim}, we operate on whitened and normalized embeddings $\mathcal{U} = \{u_1, \dots, u_N\} \subset \mathbb{S}^{d-1}$. Since CLIP encodes semantics directionally and compares representations via cosine similarity, the natural probability model on $\mathbb{S}^{d-1}$ is the von Mises–Fisher (vMF) distribution.

\noindent\textbf{Single vMF Distribution.} For a unit vector $u \in \mathbb{S}^{d-1}$, the density of a vMF distribution with mean direction $\mu \in \mathbb{S}^{d-1}$ and concentration $\kappa \ge 0$ is
\begin{equation}
    f_{\text{vMF}}(u; \mu, \kappa) = C_d(\kappa) \exp(\kappa \mu^\top u),
\end{equation}
where the normalization constant is
\begin{equation}
    C_d(\kappa) = \frac{\kappa^{d/2-1}}{(2\pi)^{d/2} I_{d/2-1}(\kappa)}.
\end{equation}
and $I_\nu(\cdot)$ denotes the modified Bessel function of the first kind. The mean direction $\mu$ identifies the center of a semantic cluster on the hypersphere, while $\kappa$ controls its concentration. Larger $\kappa$ corresponds to tighter clustering around $\mu$, whereas $\kappa=0$ reduces to the uniform distribution on $\mathbb{S}^{d-1}$.

\noindent\textbf{Mixture Model.} Natural image and text distributions are inherently multimodal, containing distinct semantic regions (\emph{e.g.,} animals, vehicles, landscapes). To capture this structure, we model the latent density as a mixture of $K$ independent vMF components:
\begin{equation}
    p(u \mid \Theta)
    =
    \sum_{k=1}^K \pi_k \,
    f_{\text{vMF}}(u; \mu_k, \kappa_k),
\end{equation}
where $\Theta = \{(\pi_k, \mu_k, \kappa_k)\}_{k=1}^K$ are the set parameters that need to be learned, and the mixing coefficients satisfy $\pi_k \ge 0$ and $\sum_{k=1}^K \pi_k = 1$. Each component corresponds to a coherent semantic mode on the hypersphere, and the resulting density naturally reflects the multimodal organization of the CLIP latent space.

\subsection{Parameter Estimation via EM}
\label{sec:em_algo}

We estimate the parameters $\Theta$ by maximizing the log-likelihood
\[
\mathcal{L}(\Theta)
=
\sum_{i=1}^N
\log p(u_i \mid \Theta),
\]
using the Expectation–Maximization (EM) algorithm for vMF distributions~\citep{JMLR:v6:banerjee05a}. The EM procedure alternates between computing the vMF posterior assignments (E-step) and updating the model parameters (M-step).

\noindent\textbf{E-Step.} Given current parameters, we compute the normalized responsibility
\[
\gamma_{ik}
=
p(k \mid u_i, \Theta)
=
\frac{
\pi_k f_{\text{vMF}}(u_i; \mu_k, \kappa_k)
}{
\sum_{j=1}^K
\pi_j f_{\text{vMF}}(u_i; \mu_j, \kappa_j)
}.
\]
For numerical stability in high dimensions, we evaluate this expression in the log-domain: $\log \gamma_{ik}\propto\log \pi_k+\kappa_k \mu_k^\top u_i+\log C_d(\kappa_k).$ These responsibilities quantify the soft assignment of each embedding to semantic components.

\noindent\textbf{M-Step.} In this step, we update the parameters to maximize the expected complete-data log-likelihood based on responsibilities computed in the E-step.
\paragraph{Mixing Coefficients and Mean Directions.}
The mixing coefficients are updated as the fraction of total mass assigned to each cluster:
\begin{equation}
    \pi_k^{\text{new}} = \frac{N_k}{N}, \quad \text{where } N_k = \sum_{i=1}^N \gamma_{ik}.
\end{equation}
Then the weighted resultant vector and updated mean for cluster $k$ are:
\begin{equation}
    r_k = \sum_{i=1}^N \gamma_{ik} u_i,\quad  \mu_k^{\text{new}} = \frac{r_k}{\|r_k\|_2}.
\end{equation}

\paragraph{Concentration Parameter Update.} Let $\bar{R}_k = \nicefrac{\|r_k\|_2}{N_k}$ denote the mean resultant length. The maximum-likelihood estimate of $\kappa_k$ involves
\[
A_d(\kappa_k) = \bar{R}_k,
\quad
\text{where}
\quad
A_d(\kappa) =
\frac{I_{d/2}(\kappa)}{I_{d/2-1}(\kappa)}.
\]
Direct inversion is computationally expensive, particularly in high-dimensional settings ($d \ge 512$). We therefore adopt the accurate approximation from \citet{JMLR:v6:banerjee05a}:
\begin{equation}
    \kappa_k^{\text{new}}
    \approx
    \frac{\bar{R}_k (d - \bar{R}_k^2)}
         {1 - \bar{R}_k^2}.
\end{equation}
This approximation yields stable and efficient updates suitable for large-scale CLIP embeddings while maintaining sufficient accuracy.

\subsection{Inference: Density Evaluation and Semantic Decomposition}
\label{sec:inference}

After estimating the movMF parameters $\Theta = \{(\pi_k, \mu_k, \kappa_k)\}_{k=1}^K$ via EM, the learned model defines an explicit probability density on the hypersphere. Inference therefore reduces to evaluating this density and its associated posterior structure. Importantly, since CLIP maps both images and text into a shared embedding space, the same probabilistic framework applies to either modality.

\noindent\textbf{Likelihood as a Geometric Density Score.} For a query sample $u \in \mathbb{S}^{d-1}$, the log-likelihood score $S(u)$ under the learned mixture is
\begin{equation}
    S(u) = \log P(u | \Theta) = \log \left( \sum_{k=1}^K \pi_k C_d(\kappa_k) \exp(\kappa_k \mu_k^\top u) \right).
\end{equation}
This quantity measures how well the sample aligns with the learned semantic modes of the latent space. Unlike Gaussian-based scoring, the likelihood here reflects proximity to multiple semantic directions rather than distance from a single global center. Since the model captures multimodal density structure, low likelihood indicates that a sample lies outside established semantic regions rather than merely being far from an average embedding. \textit{This enables principled evaluation of representation quality, distribution shift in image and text domains, and long-tail behavior within a unified probabilistic framework.}


\noindent\textbf{Posterior Structure and Semantic Attribution.}
\label{sec:interpretability}
Beyond scalar likelihood scoring, the mixture structure provides a richer, interpretable decomposition. For a given embedding $u$, we can compute the posterior responsibility
\begin{equation}
\gamma_k(u)
=
P(k \mid u)
=
\frac{
\pi_k f_{\text{vMF}}(u; \mu_k, \kappa_k)
}{
\sum_{j=1}^K 
\pi_j f_{\text{vMF}}(u; \mu_j, \kappa_j)
}.
\label{eq:posterior}
\end{equation}
This posterior distribution quantifies the degree to which the sample is explained by each semantic component. Each mixture component corresponds to a coherent directional mode on the hypersphere. By associating mixture centers with representative samples or textual descriptions from the training data, these components can be interpreted as semantic prototypes. The posterior vector $\{\gamma_k(u)\}_{k=1}^K$ therefore provides a soft semantic decomposition of the embedding:
\[
u \;\approx\; \sum_{k=1}^K \gamma_k(u)\, \mu_k,
\]
where we can interpret the representation as a weighted combination of latent semantic modes. This interpretation arises directly from the probabilistic structure of the model and does not require auxiliary decoders or external supervision. Our MovMF-CLIP framework yields both \textit{\textbf{accurate density estimation}} and \textit{\textbf{intrinsic semantic attribution}} within a unified hyperspherical geometry.

\section{Experiments}

\subsection{Experimental Setup}


\noindent\textbf{Datasets.} We adopt MS-COCO 2017 as the primary in-distribution (ID) dataset. Its validation split is used for both density estimation and long-tailed evaluation across semantic categories. For OOD evaluation, we use a cleaned subset of OpenImages~\citep{Kuznetsova_2020}, following VOS~\citep{du2022voslearningdontknow} to remove overlapping categories.

\noindent\textbf{Long-tailed and OOD Detection.} We assess likelihood robustness by measuring how well the learned density distinguishes ID samples from OOD data and how it behaves across frequent versus rare semantic categories. This setup allows us to evaluate whether hyperspherical mixture modeling mitigates the common conflation between semantic rarity and distributional abnormality.

\noindent\textbf{Interpretable Semantic Decomposition.} To evaluate interpretability, we compare against SPLICE~\citep{bhalla2024interpretingclipsparselinear}, MSAE~\citep{zaigrajew2025interpretingcliphierarchicalsparse}, and TextSpan~\citep{gandelsman2024interpretingclipsimagerepresentation} by measuring alignment between mixture-based semantic decomposition and ground-truth captions. This metric evaluates whether the posterior responsibilities of mixture components correspond to meaningful linguistic concepts in the annotations.

\noindent\textbf{Semantic Stability Under Iterative Generative Drift.} To study generative robustness, we simulate semantic drift through iterative encoding–decoding cycles using the pipeline of CLIP + UnCLIP\footnote{https://huggingface.co/sd2-community/stable-diffusion-2-1-unclip-small}. At each iteration, images are re-encoded into CLIP space, allowing us to track density and semantic changes. We compare against W-CLIP~\citep{betser2025whitenedcliplikelihoodsurrogate} in terms of semantic stability.


\noindent\textbf{Implementation Details.} For experiments on semantic decomposition, following \citet{bhalla2024interpretingclipsparselinear}, all methods use the same OpenCLIP ViT-B/32 backbone trained on LAION-400M. By contrast, the pre-trained CLIP ViT-L/14 backbone is used for all the other experiments. Following \citet{betser2025whitenedcliplikelihoodsurrogate}, we estimate the whitening transformation from MS-COCO validation set. The hyperparameters of our method are fixed across tasks. Additional qualitative analyses and extended results on BLIP-2~\citep{li2023blip2bootstrappinglanguageimagepretraining} and CoCa~\citep{yu2022cocacontrastivecaptionersimagetext}  are provided in Appendices~C and E.


\begin{table}[t]
\centering
\caption{\textbf{Long-tail likelihood fairness.} AUROC for separating head and tail categories using likelihood scores (ideal $\rightarrow 0.5$). Lower values indicate reduced bias against rare concepts. MovMF-CLIP substantially improves fairness compared to W-CLIP.}
\label{tab:long_tail_fairness}
\begin{tabular}{@{}lcc@{}}
\toprule
\textbf{Method} & W-CLIP~\citep{betser2025whitenedcliplikelihoodsurrogate}   &MovMF-CLIP (Ours) \\
\midrule
\textbf{AUROC}  & 0.7826   & \textbf{0.5819} \\
\bottomrule
\end{tabular}
\end{table}

\begin{table}[t]
\centering
\caption{\textbf{Out-of-distribution detection.} FPR95 ($\downarrow$) and AUROC ($\uparrow$) on MS-COCO (full set and tail-only subset) versus OpenImages OOD samples. MovMF-CLIP improves both detection performance and maintains robustness on long-tail subsets.}
\label{tab:ood_detection}
\setlength{\tabcolsep}{3.5mm}
\resizebox{0.99\linewidth}{!}{
\begin{tabular}{@{}lcccc@{}}
\toprule
\multirow{2}{*}{\textbf{Method}} & \multicolumn{2}{c}{\textbf{ID: Full COCO}} & \multicolumn{2}{c}{\textbf{ID: Tail Subset Only}} \\
\cmidrule(lr){2-3} \cmidrule(lr){4-5}
& FPR95 ($\downarrow$) & AUROC ($\uparrow$) & FPR95 ($\downarrow$) & AUROC ($\uparrow$) \\
\midrule
MCM~\citep{ming2022delving} & 84.50\% & 0.8204 & 62.90\% & 0.8754 \\
EOE~\citep{cao2024envisioningoutlierexposurelarge} & 83.91\% & 0.7926 & 75.54\% & 0.8364 \\
NegLabel~\citep{jiang2024negativelabelguidedood} & 56.48\% & 0.8564 & 38.77\% & 0.9135 \\
W-CLIP~\citep{betser2025whitenedcliplikelihoodsurrogate}           & 67.76\% & 0.8600 & 75.05\% & 0.7842 \\
\midrule
\textbf{MovMF-CLIP (Ours)}   & \textbf{48.00\%} & \textbf{0.8952} & \textbf{33.48\%} & \textbf{0.9157} \\
\bottomrule
\end{tabular}
}
\end{table}

\subsection{Long-tailed and OOD Detection}
\label{sec:ood_detection}

A key limitation of unimodal latent density modeling (\emph{e.g.}, W-CLIP) is the conflation of \textbf{\emph{semantic rarity}} with \textbf{\emph{distributional abnormality}}. Under a single isotropic Gaussian assumption, valid but infrequent long-tail concepts tend to receive low likelihood simply due to their distance from the global mean, making them indistinguishable from true OOD samples. In contrast, MovMF-CLIP models the latent space as a hyperspherical mixture, allowing rare but semantically coherent regions to be represented by dedicated mixture components. We evaluate this property through long-tail likelihood fairness and OOD detection.

\noindent\textbf{Long-Tailed Likelihood Fairness (texts).} We analyze captions of MS-COCO, partitioning into \emph{head} and \emph{tail} groups based on concept frequency statistics. To quantify bias against rare semantics, we measure how well likelihood scores separate head from tail samples using AUROC. A value closer to $0.5$ indicates fairness, meaning likelihood does not systematically penalize rare concepts.

As shown in Table~\ref{tab:long_tail_fairness}, W-CLIP exhibits strong discrimination against tail samples (AUROC $> 0.78$), confirming that a global Gaussian prior treats long-tail concepts as outliers. In contrast, MovMF-CLIP substantially reduces this bias (AUROC closer to $0.5$), indicating that multimodal hyperspherical modeling assigns calibrated likelihoods across semantic frequency groups.

\noindent\textbf{Out-of-Distribution Detection (images).} 
Table~\ref{tab:ood_detection} shows that MovMF-CLIP consistently outperforms all competing approaches. On the full ID set, FPR95 is reduced from $67.76\%$ to $48.00\%$ compared with W-CLIP. When restricting ID samples to the pure tail subset, the degradation observed in W-CLIP becomes pronounced (FPR95 increases from $67.76\%$ to $75.05\%$), while MovMF-CLIP remains stable and improves FPR95 to $33.48\%$. This demonstrates that separating semantic multimodality from global density structure improves both fairness to long-tail concepts and robustness to distributional shifts. Consistent improvements are also observed on BLIP-2~\citep{li2023blip2bootstrappinglanguageimagepretraining}/CoCa~\citep{yu2022cocacontrastivecaptionersimagetext} backbones (see Appendix~C). 

\begin{table}[t]
\centering
\caption{\textbf{Semantic Relevance and inference efficiency.} Comparison with concept-based decomposition methods using the same OpenCLIP ViT-B/32 backbone. MovMF-CLIP achieves higher semantic alignment with ground-truth captions while reducing per-image inference time through its closed-form posterior computation.}
\label{tab:splice_comparison}
\renewcommand{\arraystretch}{1.1}
\resizebox{0.99\linewidth}{!}{
\begin{tabular}{lcc}
\toprule
\textbf{Method} & \textbf{Semantic Relevance ($\uparrow$)} & \textbf{Inference Time (ms) ($\downarrow$)} \\
\midrule
MSAE~\citep{zaigrajew2025interpretingcliphierarchicalsparse}                      & 0.530           & 14.6\\
TextSpan~\citep{gandelsman2024interpretingclipsimagerepresentation}                      & 0.538           & 29.4\\
CLIP + Sparse Decomposition~\citep{bhalla2024interpretingclipsparselinear} & 0.585           & 112.2 \\
Negative Concept Weights~\citep{bhalla2024interpretingclipsparselinear}    & 0.635           & 102.8\\
SPLICE~\citep{bhalla2024interpretingclipsparselinear}                      & 0.655           & 132.7\\
\midrule
\textbf{MovMF-CLIP (Ours)}  & \textbf{0.673} & \textbf{9.8} \\
\bottomrule
\end{tabular}
}
\end{table}

\subsection{Interpretable Semantic Decomposition}
\label{sec:interpretable_decomposition}

We evaluate whether the learned mixture structure learned yields semantically meaningful and human-aligned decompositions of visual representations. 

\noindent\textbf{Evaluation Metric.} Following SPLICE, we adopt the \textbf{Semantic Relevance} metric to quantify alignment between model-derived concepts and ground-truth captions. Higher values indicate stronger semantic agreement. Detailed formulations are kindly referred to Appendix C.

\noindent\textbf{Experimental Setup.}  For all projection-based baselines, we report results using their officially recommended optimal sparsity settings. As for our MovMF-CLIP, we use $K=500$ mixture components and retain the top $N=10$ keywords per component. Ablation studies over $K$ and $N$ are provided in Appendix~D.

\noindent\textbf{Results and Efficiency.} Table~\ref{tab:splice_comparison} shows that MovMF-CLIP achieves the highest Semantic Relevance score, outperforming projection-based baselines. Beyond accuracy, MovMF-CLIP offers a substantial computational advantage. Compared to the second best method, MovMF-CLIP significantly reduces per-image inference time from 132.7\,ms to \textbf{9.8\,ms}, achieving over a \textbf{13$\times$ speedup}. This is because projection-based methods require solving a sparse regression problem over large concept dictionaries for each image, typically involving iterative CPU-based solvers. In contrast, semantic decomposition in MovMF-CLIP reduces to evaluating posterior responsibilities $\gamma_k(u) = P(k \mid u)$ which requires only a matrix multiplication with the learned cluster centers.

\begin{wrapfigure}{r}{0.5\textwidth}
    \vspace{-3mm} 
    \centering
    \includegraphics[width=\linewidth]{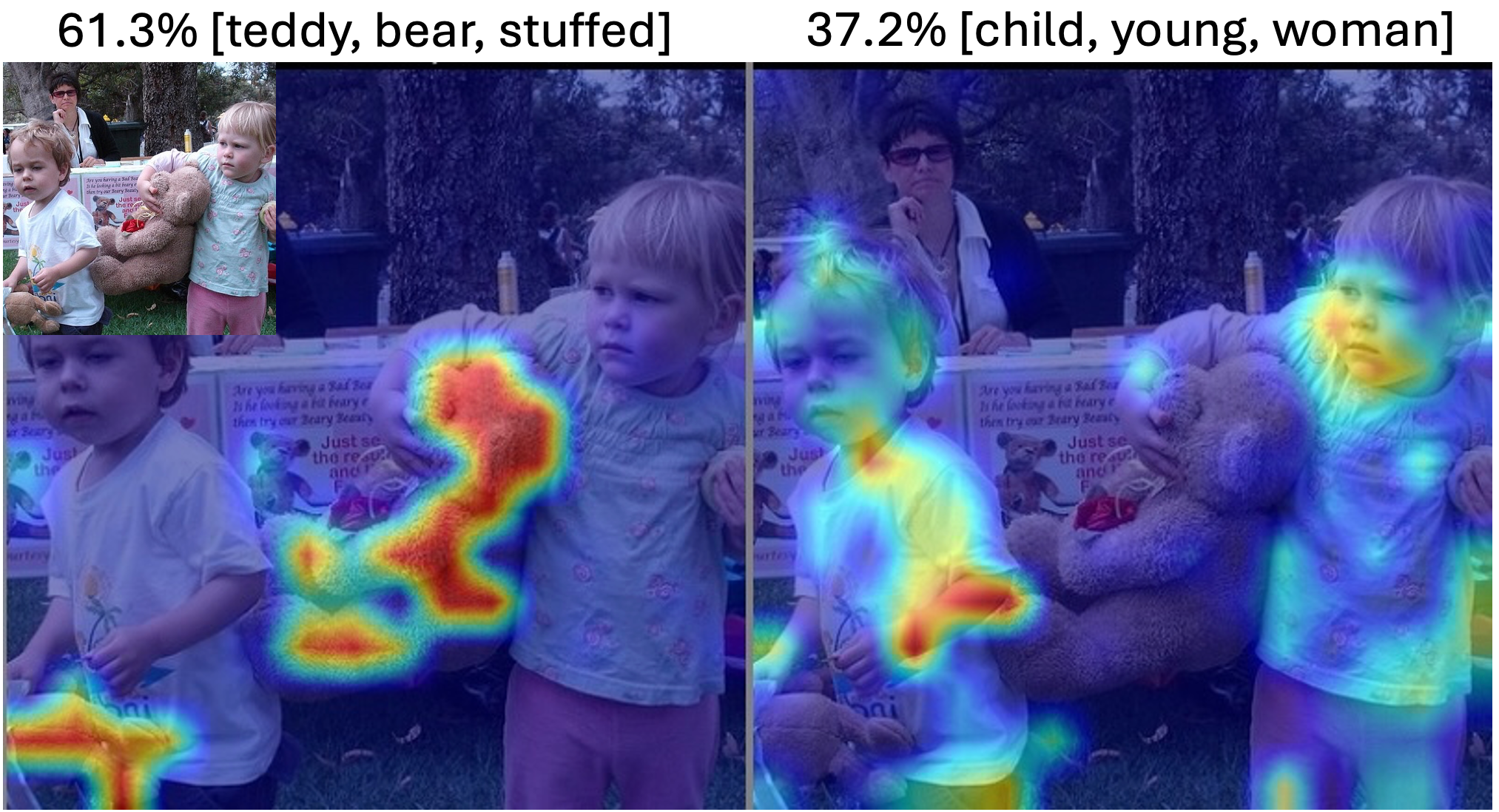}
    \caption{\textbf{Concept localization.}  MovMF-CLIP represents each image as a sparse mixture of semantic prototypes. Concept-specific heatmaps highlight image regions associated with each semantic component.}
    \label{fig:reli}
    \vspace{-3mm} 
\end{wrapfigure}

\begin{figure*}[t]
    \centering
    \includegraphics[width=0.99\linewidth]{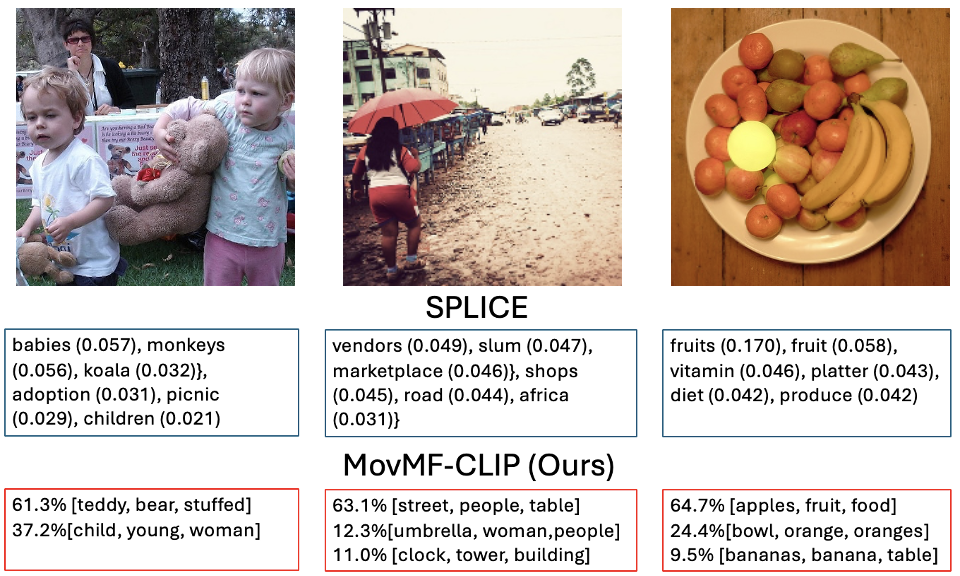} 
    \caption{\textbf{Qualitative comparison of semantic decomposition.} Concept extraction results for SPLICE and MovMF-CLIP on representative scenes. Projection-based methods may emphasize generic or context-dependent terms, whereas our MovMF-CLIP produces a sparse probabilistic combination of semantically coherent cluster prototypes aligned with object-centric content.}
    \vspace{-3mm}
    \label{fig:qualitative_case_study}
\end{figure*}

\noindent\textbf{Qualitative Analysis.} To provide an intuitive illustration of the learned semantic structure, we visualize and compare the decompositions produced by MovMF-CLIP and SPLICE on several complex real-world scenes (see \cref{fig:qualitative_case_study}). Projection-based methods often yield broad or loosely related concepts when representing images as sparse combinations over large dictionaries. For example, in the fruit bowl scene, SPLICE primarily identifies generic terms such as “fruit” or “vitamin”, without resolving more specific semantics. Our MovMF-CLIP decomposes each image into a sparse probabilistic combination of hyperspherical semantic prototypes. The resulting components tend to align with more precise and fine-grained visual concepts in the scene. This behavior is consistent with the multimodal structure induced by hyperspherical mixture modeling. Fig.~\ref{fig:reli} further visualizes the concept-specific heatmaps produced by our model, where the highlighted regions closely correspond to the underlying objects associated with each semantic component.

\begin{figure*}[t]
    \centering
    \includegraphics[width=0.99\linewidth]{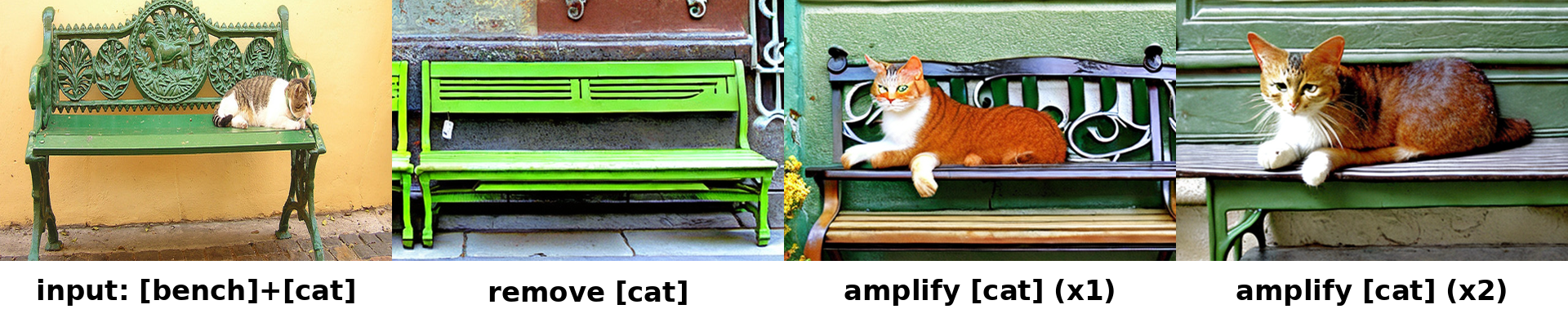}
    \caption{\textbf{Directional semantic intervention in CLIP latent space.} We identify a vMF component associated with the cat concept and directly manipulate its direction in the whitened CLIP space before UnCLIP decoding. Removing the component suppresses the cat from the generated image, while amplifying the same direction strengthens the corresponding visual semantics. This illustrates that MovMF-CLIP learns semantically meaningful and controllable directions.}
    \label{fig:semantic_direction_editing}
\end{figure*}

\noindent\textbf{Semantic Direction Intervention.}
Beyond decomposing embeddings into interpretable concepts, the learned vMF components can also be used as explicit semantic directions. Given a target component center $\mu_{k}$, we manipulate the whitened embedding by subtracting or amplifying its projection along $\mu_{k}$, while keeping the overall embedding norm fixed before mapping it back for UnCLIP decoding. As shown in Fig.~\ref{fig:semantic_direction_editing}, removing the cat-associated direction suppresses the cat semantics, whereas amplifying the same direction progressively strengthens them. This provides qualitative evidence that the mixture components capture controllable semantic factors rather than merely improving aggregate likelihood scores.

\FloatBarrier

\subsection{Semantic Stability Under Iterative Generative Drift}
\label{sec:unclip_generation}
We simulate a process of iterative generative drift by constructing the encoding–decoding pipeline in Fig.~\ref{fig:pipeline4}, which involves conducting CLIP encoding and UnCLIP decoding repeatedly. When an image is projected into the latent space and decoded back to the pixel domain iteratively, small perturbations accumulate, gradually shifting the embedding toward low-density regions between valid semantic modes. This \emph{semantic drift} manifests as progressive degradation in both visual fidelity and semantic consistency. 
We evaluate whether the hyperspherical mixture structure learned by MovMF-CLIP can serve as a geometric prior to stabilize CLIP latent representations during such iterative processes.

\begin{center}
    \begin{minipage}{0.78\linewidth}
    \centering
    \includegraphics[width=\linewidth]{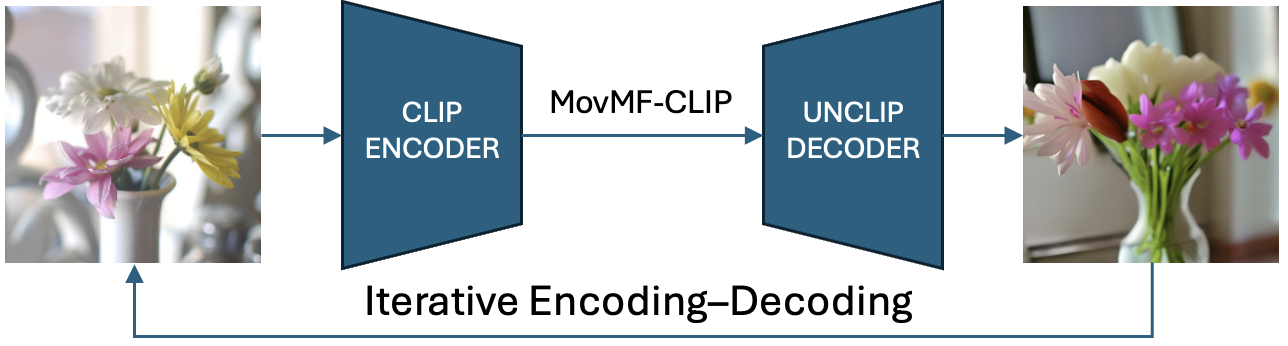}
    \captionof{figure}{\textbf{Pipeline for iterative generative drift analysis.} We insert MovMF-CLIP into the encoding-decoding loop to stabilize the underlying semantics.}
    \label{fig:pipeline4}
    \end{minipage}
    \vspace{-3mm}
\end{center}

\noindent\textbf{Semantic Stabilization via Hyperspherical Mixture Projection.}
Given a whitened embedding $\tilde{z}$, we first normalize it onto the hypersphere and compute its posterior responsibilities $\{\gamma_k\}$. Components with negligible posterior mass are discarded, and the embedding is reconstructed as a weighted combination of the retained mixture centers. The reconstructed direction is then rescaled to match the original whitened norm and mapped back to the original space via inverse whitening. This operation projects drifting embeddings toward high-density semantic regions while preserving their magnitudes. The resulting embedding is then fed into the next UnCLIP decoding step.

\noindent\textbf{Experimental Setup.} We mainly compare MovMF-CLIP against W-CLIP~\citep{betser2025whitenedcliplikelihoodsurrogate}, which regularizes embeddings by projecting them onto a fixed-radius hypersphere in whitened space without directional correction. To quantify semantic stability, we measure two metrics between the original image and the image obtained after six iterations: LPIPS (AlexNet)~\citep{zhang2018unreasonableeffectivenessdeepfeatures} for perceptual distortion ($\downarrow$ better) and CLIP cosine similarity for semantic retention ($\uparrow$ better).


\noindent\textbf{Quantitative Results.} As shown in Table~\ref{tab:unclip_metrics}, MovMF-CLIP consistently improves both perceptual quality and semantic stability. While W-CLIP constrains embedding magnitude, it does not regulate directional drift on the hypersphere, leading to gradual semantic degradation. In contrast, our mixture-based method explicitly anchors embeddings to high-density semantic modes, resulting in improved cosine similarity and reduced perceptual distortion.

\begin{center}
    \begin{minipage}{\linewidth}
    \captionof{table}{\textbf{Iterative CLIP--UnCLIP generation.} Average LPIPS ($\downarrow$) and CLIP cosine similarity ($\uparrow$) over $1,000$ MS-COCO images after $k=6$ encode--decode iterations. MovMF-CLIP improves both perceptual consistency and semantic retention.}
    \label{tab:unclip_metrics}
    \centering
    \setlength{\tabcolsep}{4mm}
    \begin{tabular}{lcc}
    \toprule
    \textbf{Method} & \textbf{LPIPS} ($\downarrow$) & \textbf{CLIP Cosine} ($\uparrow$) \\
    \midrule
    Vanilla CLIP-UnCLIP & 	0.7733 & 	0.5911 \\
    W-CLIP~\citep{betser2025whitenedcliplikelihoodsurrogate} & 0.7678 & 0.6171 \\
    MovMF-CLIP (Ours)  & \textbf{0.7357} & \textbf{0.6710} \\
    \bottomrule
    \end{tabular}
    \end{minipage}
\end{center}

\noindent\textbf{Qualitative Analysis.} Fig.~\ref{fig:unclip_case_studies} visualizes representative iterative trajectories. For W-CLIP, embeddings progressively drift away from coherent semantic regions, leading to noticeable degradation. Instead, our MovMF-CLIP maintains alignment with semantically meaningful prototypes throughout the iterations, yielding more stable visual and semantic evolution.

\begin{center}
    \begin{minipage}{0.92\linewidth}
    \centering
    \includegraphics[width=\linewidth]{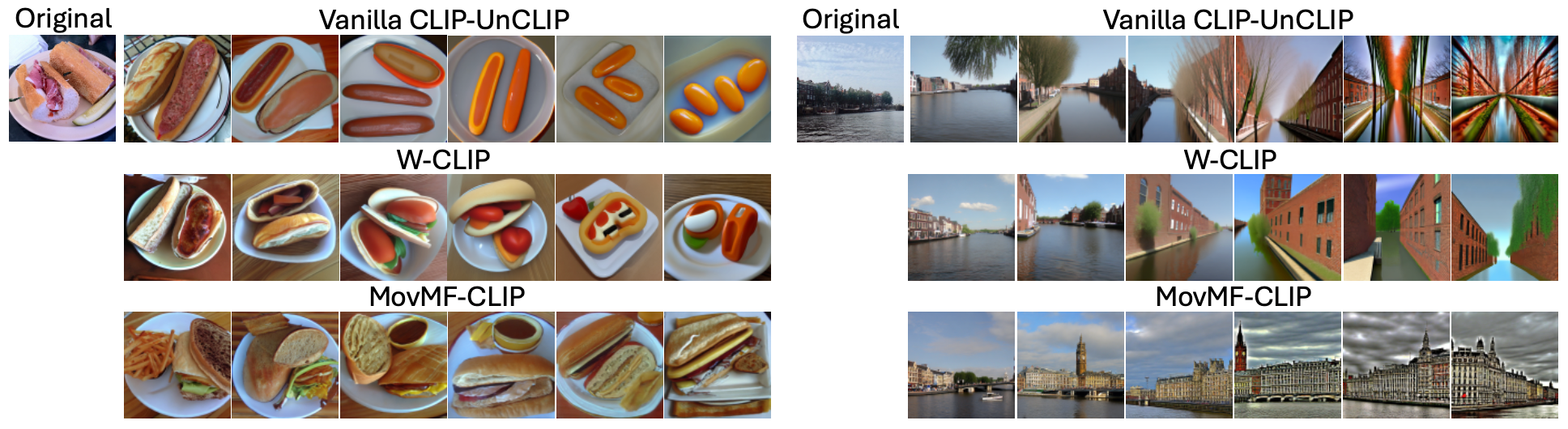}
    \captionof{figure}{\textbf{Qualitative comparison under iterative generative drift.} For each example, the original image is followed by its evolution over six encoding-decoding iterations. The three rows demonstrate the trajectories using vanilla CLIP-UnCLIP (top), W-CLIP (middle), and MovMF-CLIP (bottom). Our hyperspherical mixture aligns well with coherent semantic prototypes, resulting in more stable semantic trajectories.}
    \label{fig:unclip_case_studies}
    \end{minipage}
\end{center}

\FloatBarrier

\section{Conclusion}

We revisited CLIP latent space through a geometric lens and argued that it is more faithfully modeled as a hyperspherical semantic mixture rather than a unimodal Gaussian. By separating global covariance anisotropy from directional multimodality, MovMF-CLIP provides a simple and principled density model aligned with the intrinsic structure of contrastive embeddings. This unified geometric–probabilistic framework enables calibrated likelihood estimation, intrinsic semantic decomposition, and improved semantic stability under iterative generative drift. Our findings underscore the central role of representation geometry in probabilistic modeling of foundation models and point toward structured, geometry-aware probabilistic frameworks for future multimodal learning.

\bibliography{main}
\bibliographystyle{iclr2026_conference}

\appendix
\section{Future Work}
\label{sec:future_work}
\subsection{Connection with Score-based Generative Models}
\label{subsec:sgm}

Beyond density estimation, the hyperspherical density learned by MovMF-CLIP may provide a useful geometric prior for generative modeling. Modern latent diffusion and score-based models~\citep{song2021scorebasedgenerativemodelingstochastic,rombach2022highresolutionimagesynthesislatent,vahdat2021scorebasedgenerativemodelinglatent} typically assume simple Gaussian priors in the latent space. In contrast, the movMF mixture defines a directional density on the hypersphere, which naturally yields a closed-form score function on the manifold. This observation suggests a potential connection between hyperspherical density modeling and Riemannian score-based generative models~\citep{debortoli2022riemannianscorebasedgenerativemodelling,huang2022riemanniandiffusionmodels}. In particular, the learned density could provide a geometry-aware guidance signal that encourages latent trajectories to remain within semantically meaningful regions of the representation space. Exploring such integrations with latent diffusion or unCLIP-style generation pipelines is an interesting direction for future work.

\subsection{Hierarchical Modeling of Semantic Structure}
\label{subsec:hierarchical}

Our MovMF-CLIP currently models CLIP latent space using a flat mixture with a fixed number of components. While this formulation already captures multimodal structure, semantic concepts in natural data often exhibit hierarchical organization. A natural extension is therefore to explore hierarchical mixture formulations that allow components to be organized across multiple semantic levels. For example, Bayesian nonparametric approaches such as hierarchical Dirichlet processes could allow the number of semantic components to adapt to the data automatically. Such hierarchical modeling may provide a coarse-to-fine representation of semantic structure and could potentially improve robustness to long-tail concepts by allowing rare categories to share statistical strength with related parent clusters. Investigating hierarchical extensions of our hyperspherical mixture models remains an important and interesting future direction.

\section{Why Whitening \emph{Before} Hyperspherical Mixture Modeling?}
\label{sec:whitening}

CLIP embeddings are compared by cosine similarity, which motivates modeling their directions on a hypersphere. However, the raw latent distribution is typically far from isotropic: second-order anisotropy introduces dominant covariance directions that can distort angular geometry and confound hyperspherical clustering. Let $z \in \mathbb{R}^d$ denote CLIP embeddings with mean $\mu$ and covariance
\[
\Sigma \;=\; \mathbb{E}\!\left[(z-\mu)(z-\mu)^\top\right].
\]
Empirically, $\Sigma \neq I$, reflecting global variations induced by dataset bias, frequent concepts, and representation artifacts.

\noindent\textbf{Whitening as metric normalization.}
We first apply a whitening transform in $\mathbb{R}^d$:
\[
y \;=\; W (z-\mu), \qquad W \approx \Sigma^{-1/2},
\]
so that $y$ has approximately standardized second-order statistics:
\[
\mathbb{E}[y]\approx 0, \qquad \mathrm{Cov}(y)\approx I.
\]
This step removes nuisance anisotropy and defines an \textbf{intrinsic Mahalanobis geometry} in the original space. In particular, the angular similarity in whitened space corresponds to a Mahalanobis cosine similarity in the original coordinates:
\[
\frac{y_i^\top y_j}{\|y_i\|\|y_j\|}
\;=\;
\frac{(z_i-\mu)^\top \Sigma^{-1} (z_j-\mu)}
{\|z_i-\mu\|_{\Sigma^{-1}}\,\|z_j-\mu\|_{\Sigma^{-1}}}.
\]
Thus, whitening calibrates directions according to the covariance structure of the latent distribution rather than letting a few high-variance directions dominate. This ensures that the resultant hypersphere is defined under the intrinsic second-order geometry of the data.

\noindent\textbf{Normalization and hyperspherical density.}
After whitening, we normalize to the unit hypersphere:
\[
u \;=\; \frac{y}{\|y\|} \in \mathbb{S}^{d-1},
\]
and model the directional density with a mixture of von Mises--Fisher distributions:
\[
p(u)
\;=\;
\sum_{k=1}^K \pi_k\, C_d(\kappa_k)\,\exp\!\big(\kappa_k\, \mu_k^\top u\big),
\qquad \|\mu_k\|=1.
\]

\noindent\textbf{Why not fit movMF directly on $z/\|z\|$?}
If we normalize first, $u_0 = z/\|z\|$, the subsequent movMF fit must simultaneously explain (i) global anisotropic scaling and (ii) genuine semantic multimodality. In practice, this causes mixture components to align with dominant covariance directions rather than semantic modes, leading to redundant components, unstable concentration estimates, and degraded likelihood calibration. Whitening \emph{before} normalization removes global second-order effects in $\mathbb{R}^d$ and yields a hyperspherical representation whose angular geometry better reflects true density structure.

\noindent\textbf{Importantly, whitening does not impose Gaussianity.}
Unlike approaches that use whitening to justify a unimodal Gaussian likelihood, we use whitening only to normalize second-order geometry. The distribution on the hypersphere remains explicitly multimodal and is captured by the vMF mixture. This separation between global covariance normalization and multimodal semantic density improves identifiability, statistical efficiency of EM, and interpretability of mixture components as semantic prototypes.


\section{Extended Experimental Results}
\label{sec:appendix_other_vlms}

\subsection{Implementation Details.}
\label{subsec:imp_details}
\noindent\textbf{Details of the Semantic Relevance Metric.} For a given image, let $W_{\text{cluster}}$ denote the set of keywords extracted from the most activated mixture components, and let $W_{\text{caption}}$ denote the content words (nouns and verbs) in the ground-truth caption. We compute the Hausdorff distance $d_H$ between the two embedding sets and define
\begin{equation}
\text{Semantic Relevance}
=
1 - d_H\big(E(W_{\text{cluster}}), E(W_{\text{caption}})\big),
\end{equation}
where $E(\cdot)$ denotes the text encoder. Higher values indicate stronger semantic agreement between the decomposed representation and the annotations.

\noindent\textbf{Inference Measurement Details.} All inference speeds are measured as the average per-image inference time on a workstation equipped with a single NVIDIA A100 GPU and an Intel Xeon Platinum 8468 CPU.

\subsection{Long-tailed and OOD Detection for Other VLMs}
\label{subsec:long_tail_ood}
To verify that the conflation of semantic rarity and distributional abnormality is a universal geometric flaw and that our MovMF-CLIP provides a universal remedy, we extend our experiments to two other popular VLMs: BLIP-2 and CoCa. We evaluate these models using the same long-tailed fairness and OOD detection protocols established in Section 4.2.

\noindent\textbf{Long-Tailed Fairness across VLMs.} As shown in \cref{tab:supp_fairness_vlms}, the unimodal Gaussian assumption used in W-CLIP consistently exhibits bias against tail categories across different VLM architectures, producing high AUROC values when separating head and tail samples. In contrast, MovMF-CLIP substantially reduces this bias, bringing the AUROC closer to the ideal value of $0.5$. These results suggest that modeling the latent space as a hyperspherical mixture provides a more balanced density estimate across semantic frequencies, and the improvements remain consistent across models such as BLIP-2 and CoCa.

\begin{table}[ht]
\centering
\caption{\textbf{Long-tailed fairness across VLMs.} AUROC ($\to 0.5$) for distinguishing head and tail categories using likelihood scores. MovMF-CLIP consistently reduces bias against tail concepts compared with W-CLIP across BLIP-2 and CoCa.}
\label{tab:supp_fairness_vlms}
\begin{tabular}{@{}lcccc@{}}
\toprule
\textbf{Model} & \multicolumn{2}{c}{BLIP-2} & \multicolumn{2}{c}{CoCa} \\
 \cmidrule(lr){2-3} \cmidrule(l){4-5} 
\textbf{Method} & W-CLIP & MovMF-CLIP & W-CLIP & MovMF-CLIP \\
\midrule
\textbf{AUROC} & 0.6809 & \textbf{0.5172} & 0.8308 & \textbf{0.5465} \\
\bottomrule
\end{tabular}
\end{table}

\noindent\textbf{OOD Detection across VLMs.} \cref{tab:supp_ood_vlms} reports OOD detection results using the OpenImages subset as the OOD dataset. Across both BLIP-2 and CoCa, MovMF-CLIP consistently improves over W-CLIP on both the full ID set and the tail-only subset. The improvements remain pronounced on the tail subset, where unimodal Gaussian modeling tends to misclassify rare but valid samples as OOD. For example, on CoCa the FPR95 on tail samples decreases from \textbf{92.12\%} to \textbf{80.02\%}, indicating that hyperspherical mixture modeling better preserves long-tail semantics while maintaining strong OOD detection performance.

\begin{table}[ht]
\centering
\caption{\textbf{OOD detection across VLMs.} 
AUROC ($\uparrow$) and FPR95 ($\downarrow$) on BLIP-2 and CoCa using OpenImages as the OOD dataset. 
MovMF-CLIP consistently improves detection performance and significantly reduces false rejection of tail samples.}
\label{tab:supp_ood_vlms}
\setlength{\tabcolsep}{3mm}
\resizebox{\textwidth}{!}{
\begin{tabular}{@{}llcccc@{}}
\toprule
\multirow{2}{*}{\textbf{Model}} & \multirow{2}{*}{\textbf{Method}} & \multicolumn{2}{c}{\textbf{ID: Full COCO}} & \multicolumn{2}{c}{\textbf{ID: Tail Subset Only}} \\
\cmidrule(lr){3-4} \cmidrule(lr){5-6}
& & AUROC ($\uparrow$) & FPR95 ($\downarrow$) & AUROC ($\uparrow$) & FPR95 ($\downarrow$) \\
\midrule
\multirow{2}{*}{BLIP-2} 
 & W-CLIP  & 0.9352 & 32.02\% & 0.8473 & 46.65\% \\
 & MovMF-CLIP  & \textbf{0.9522} & \textbf{25.32\%} & \textbf{0.9197} & \textbf{45.57\%} \\
\midrule
\multirow{2}{*}{CoCa} 
 & W-CLIP  & 0.8195 & 86.23\% & 0.6588 & 92.12\% \\
 & MovMF-CLIP  & \textbf{0.8586} & \textbf{66.14\%} & \textbf{0.8068} & \textbf{80.02\%} \\
\bottomrule
\end{tabular}
} 
\end{table}

\subsection{Visualization of Likelihood Distributions for OOD Detection}
\label{subsec:supp_ood_histograms}

\begin{figure}[htbp]
    \centering
    \begin{subfigure}{\textwidth}
        \centering
        \includegraphics[width=0.95\linewidth]{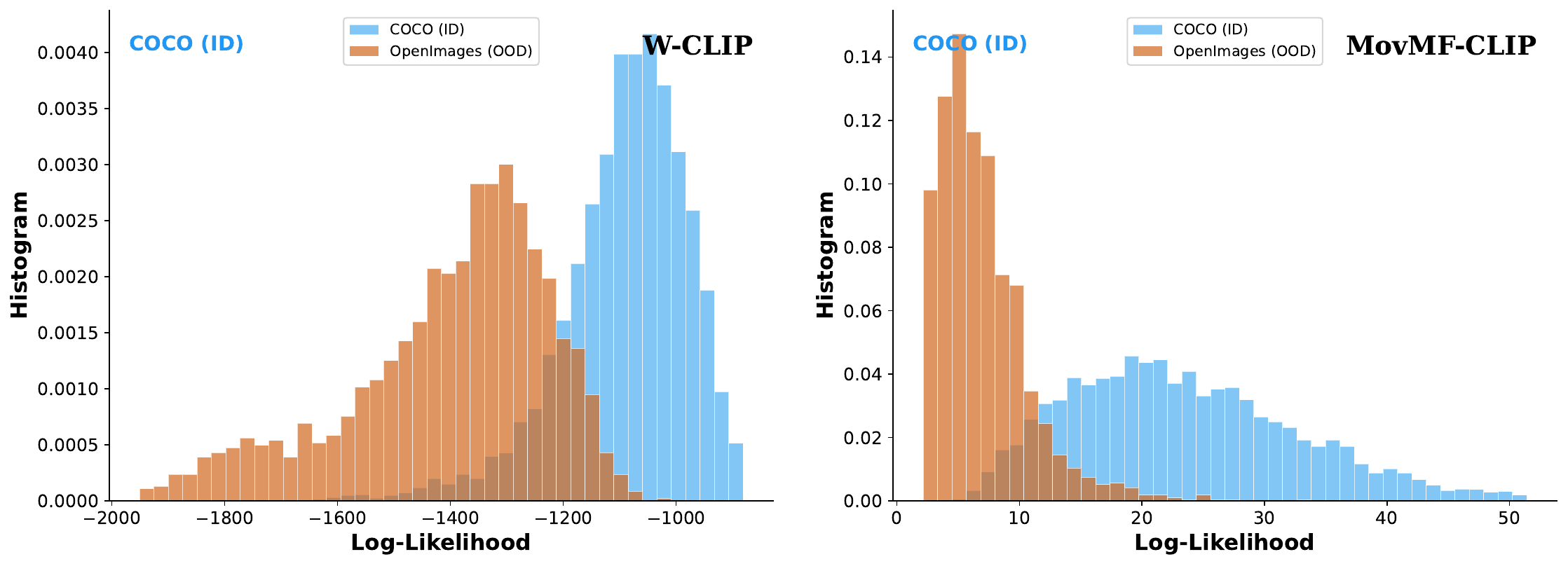}
        \caption{ID: \textbf{Full COCO} vs.\ OOD: OpenImages}
        \label{fig:ood_hist_full}
    \end{subfigure}
    
    \vspace{4mm}
    
    \begin{subfigure}{\textwidth}
        \centering
        \includegraphics[width=0.95\linewidth]{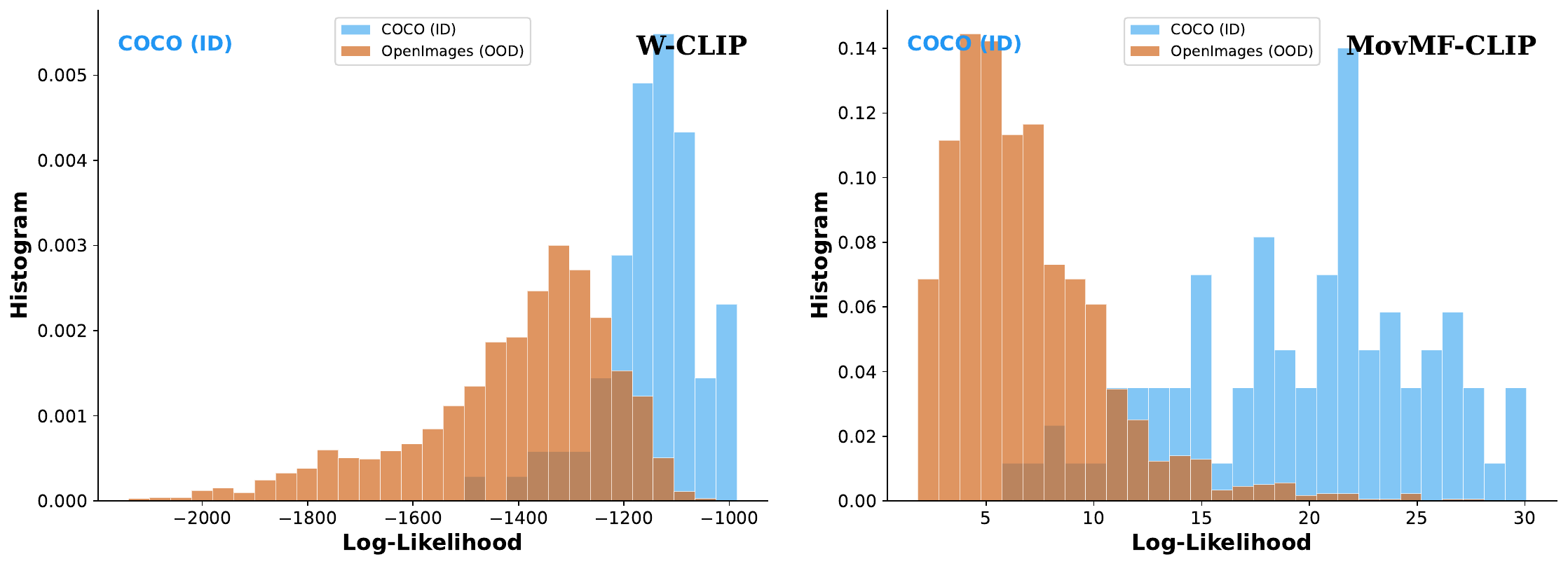}
        \caption{ID: \textbf{Tail Subset Only} vs.\ OOD: OpenImages}
        \label{fig:ood_hist_tail}
    \end{subfigure}
    
    \caption{\textbf{Likelihood distributions for OOD detection.} 
    Histograms of log-likelihood scores for ID and OOD samples. 
    \textbf{Left}: W-CLIP exhibits substantial overlap between ID and OOD distributions, particularly when the ID set contains rare concepts. 
    \textbf{Right}: MovMF-CLIP produces better separation, reflecting improved robustness to long-tail semantics while maintaining strong OOD discrimination.}
    \label{fig:supp_ood_histograms}
\end{figure}

To provide further intuition for the OOD detection results, we visualize the empirical log-likelihood distributions of ID and OOD samples in Fig.~\ref{fig:supp_ood_histograms}. The top row shows the distributions when the full MS-COCO is used as the ID dataset, while the bottom row restricts the ID samples to the tail subset containing rare concepts. In both settings, MovMF-CLIP produces a clearer separation between ID and OOD samples compared with W-CLIP. The improvement is particularly evident when the ID set contains only tail samples, where unimodal Gaussian modeling tends to assign low likelihood to rare but valid data.

\section{Ablation Studies of Hyperparameters}
\label{sec:appendix_grid_search}

In Sec.~4.3 of the main text, we showed that MovMF-CLIP outperforms the SPLICE baseline~\citep{bhalla2024interpretingclipsparselinear} in terms of Semantic Relevance. To evaluate the robustness of our approach, we conduct a grid search over two key hyperparameters: the number of mixture components ($K$) and the number of top-frequency keywords extracted per cluster ($N$).

The results are summarized in \cref{tab:movmf_relevance_grid}. Overall, MovMF-CLIP demonstrates stable performance across a broad range of configurations. Increasing the number of keywords $N$ consistently improves Semantic Relevance, as richer vocabularies better describe the semantic content of each cluster. Regarding the number of clusters $K$, performance improves when moving from small values to moderate values, reflecting the benefit of capturing more fine-grained semantic structure. When $K$ becomes very large (e.g., $K \ge 800$), the improvement saturates and slightly decreases, likely due to over-fragmentation of the latent space. The best performance is achieved with $K=500$ and $N=10$, reaching a Semantic Relevance of \textbf{0.6726}. Importantly, even small configurations (\emph{e.g.,} $K=70$, $N=3$) remain competitive, indicating that the learned hyperspherical mixture structure is robust across different sets of hyperparameters.

\begin{table}[ht]
\centering
\caption{\textbf{Grid Search for Semantic Relevance.} We report the Semantic Relevance ($1 - d_H$) on the MS-COCO validation set using OpenCLIP ViT-B/32, varying the number of clusters ($K$) and keywords per cluster ($N$).}
\label{tab:movmf_relevance_grid}
\renewcommand{\arraystretch}{1.1}
\begin{tabular}{lcccc}
\toprule
\multirow{2}{*}{\textbf{Number of Clusters ($K$)}} & \multicolumn{4}{c}{\textbf{Top-$N$ Keywords per Cluster}} \\
\cmidrule(lr){2-5}
& $N=3$ & $N=5$ & $N=7$ & $N=10$ \\
\midrule
70   & 0.6434 & 0.6552 & 0.6612 & 0.6695 \\
100  & 0.6443 & 0.6563 & 0.6624 & 0.6700 \\
300  & \textbf{0.6445} & 0.6565 & 0.6640 & 0.6722 \\
500  & 0.6442 & \textbf{0.6566} & \textbf{0.6653} & \textbf{0.6726} \\
800  & 0.6423 & 0.6561 & 0.6644 & 0.6706 \\
1000 & 0.6425 & 0.6565 & 0.6638 & 0.6711 \\
\bottomrule
\end{tabular}
\end{table}

\section{Extended Qualitative Analyses}
\label{sec:appendix_extended_analyses}

We provide extensive results to further demonstrate the robustness, fine-grained interpretability of concept localization, and semantic stability under generative drifts of our MovMF-CLIP framework.

\subsection{Semantic Decomposition Comparison}
\label{subsec:supp_decomposition}

\begin{figure*}[!ht]
    \centering
    \begin{subfigure}{\textwidth}
        \includegraphics[width=\textwidth]{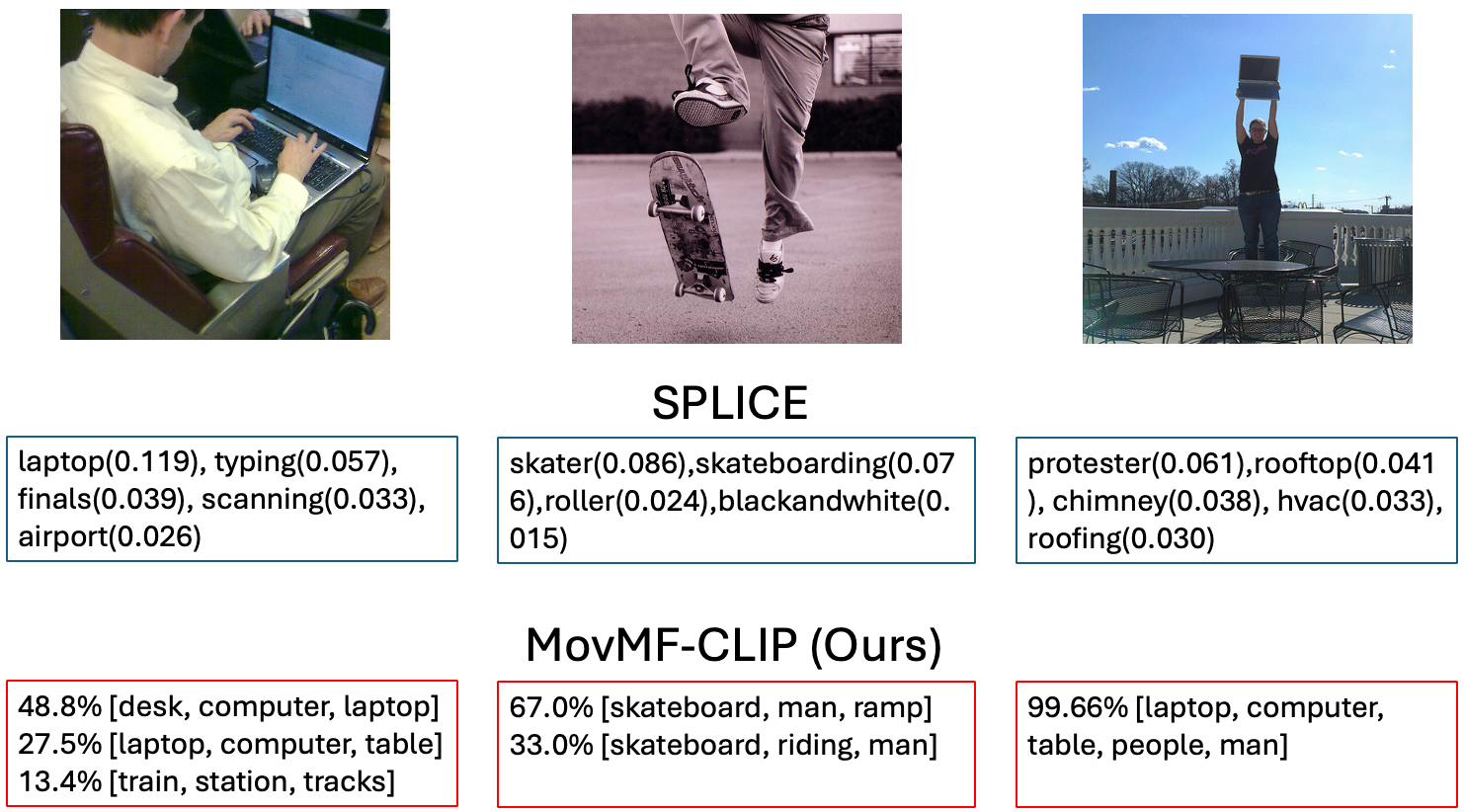} 
    \end{subfigure}
    \vspace{4mm}
    \begin{subfigure}{\textwidth}
        \includegraphics[width=\textwidth]{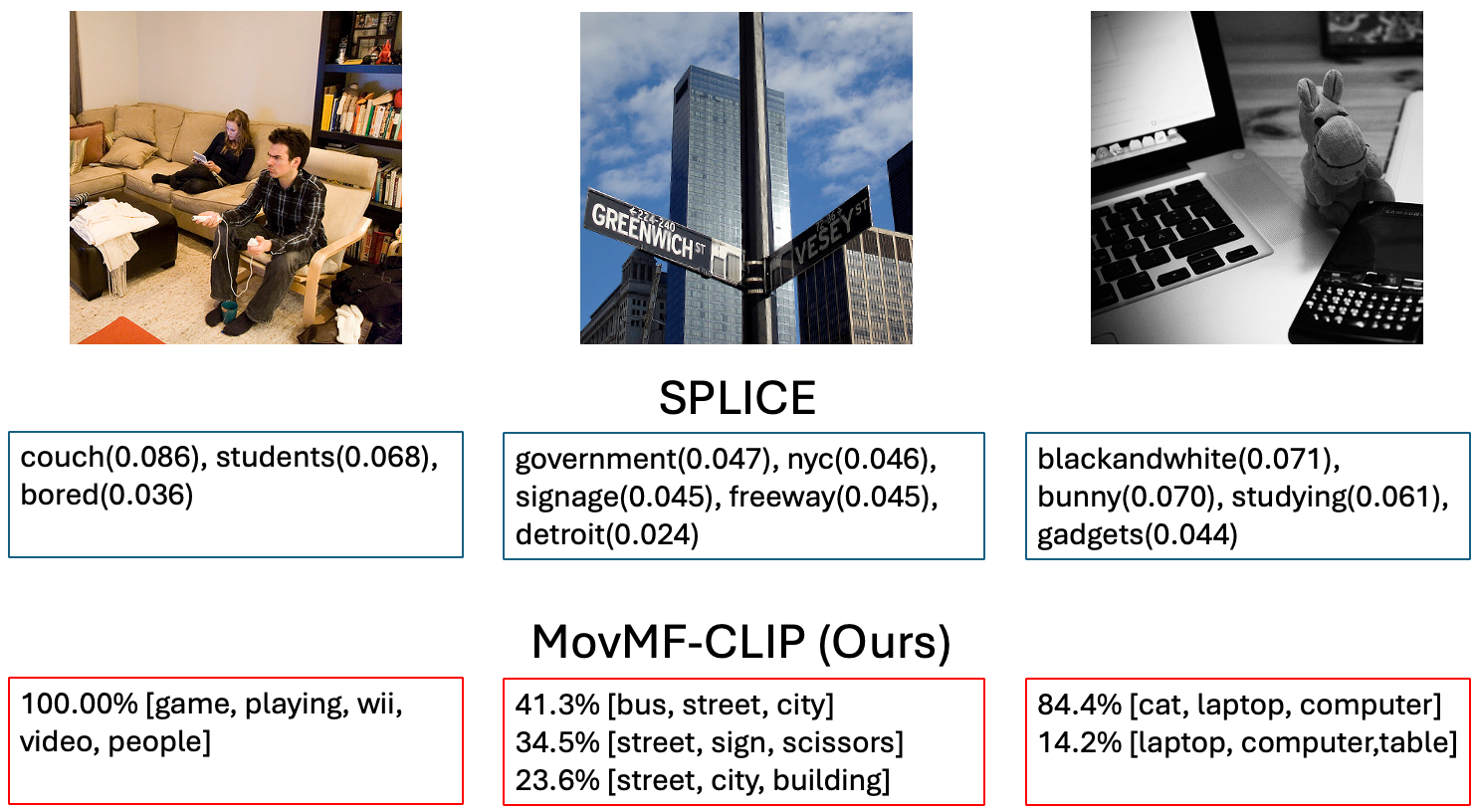} 
    \end{subfigure}
    \caption{\textbf{Extended Results of Semantic Decomposition.} Comparison of text descriptions decoded from the latent embeddings. MovMF-CLIP consistently extracts precise, fine-grained semantic concepts (e.g., ``wii'', ``skateboard''), while the baseline SPLICE often struggles with hallucinated or coarse terms (e.g., ``bored'', ``hvac'').}
    \label{fig:supp_decomposition}
\end{figure*}

To further validate our semantic decomposition capabilities (extending Fig.~4 of the main text), we present $6$ additional comparison cases against the SPLICE baseline in Fig.~\ref{fig:supp_decomposition}. 

The results clearly demonstrate that SPLICE frequently suffers from hallucinated or overly generic descriptions. For instance, given an image of a person holding a laptop (top-right), SPLICE outputs irrelevant terms like \textit{``protester''}, \textit{``chimney''}, or \textit{``hvac''}. In contrast, MovMF-CLIP precisely extracts fine-grained semantic prototypes such as \textit{``laptop''}, \textit{``table''}, and \textit{``people''}. This confirms that our hyperspherical mixture successfully maps complex visual scenes to highly accurate linguistic concepts without cross-modal contamination.

\subsection{Concept Localization}
\label{subsec:supp_heatmaps}

In Fig.~\ref{fig:supp_heatmaps}, we provide 6 additional examples of concept localization. Our MovMF-CLIP is able to successfully localize the semantic concepts. For example, in the outdoor street scene (Row 4), our method cleanly separates the \textit{``fire hydrant''} from the \textit{``dog''}. Similarly, it perfectly distinguishes a \textit{``bird''} perched on a \textit{``cow''} (Row 5), and accurately localizes specific objects in complex environments, such as isolating \textit{``wine''} from a \textit{``pizza''} on a dining table (Row 3), or a \textit{``man''} from boxes of \textit{``bananas''} (Row 6). This robust spatial grounding verifies that our learned mixture components effectively capture orthogonal semantic concepts.

\begin{figure*}[!ht]
    \centering
    \begin{subfigure}{\textwidth}
        \includegraphics[width=\linewidth]{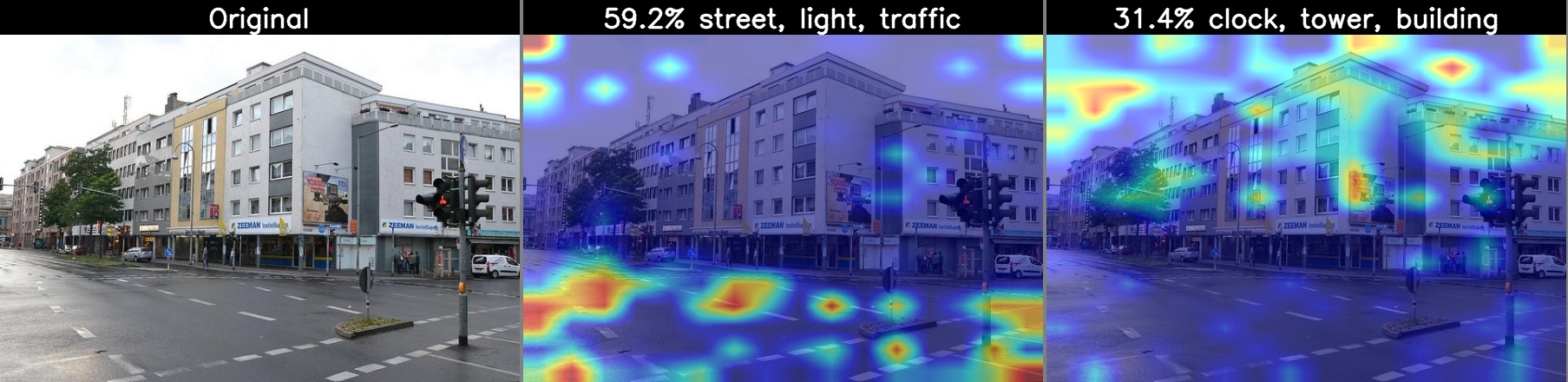}
    \end{subfigure}\vspace{1.5mm}
    \begin{subfigure}{\textwidth}
        \includegraphics[width=\linewidth]{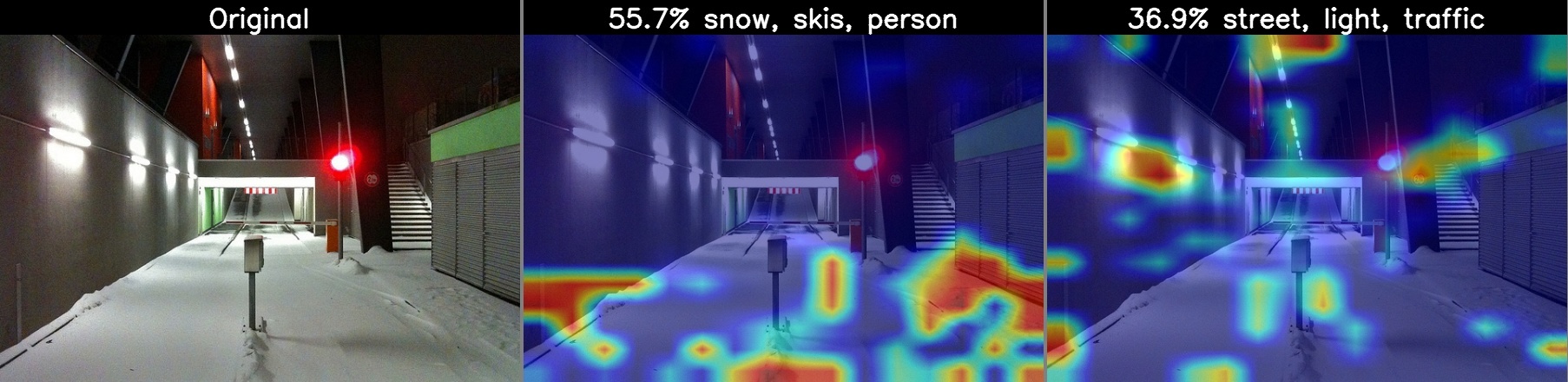}
    \end{subfigure}\vspace{1.5mm}
    \begin{subfigure}{\textwidth}
        \includegraphics[width=\linewidth]{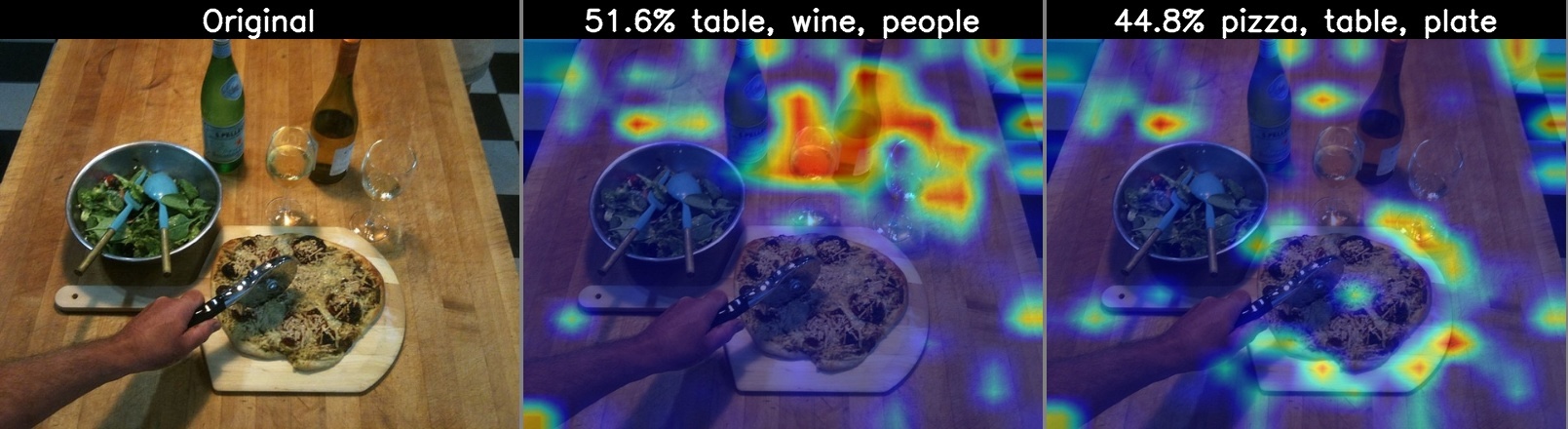}
    \end{subfigure}\vspace{1.5mm}
    \begin{subfigure}{\textwidth}
        \includegraphics[width=\linewidth]{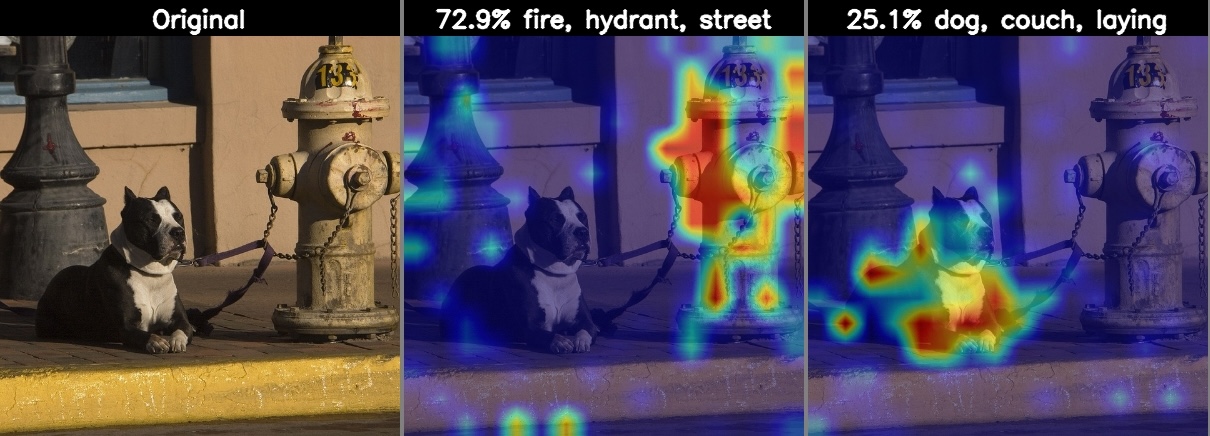}
    \end{subfigure}
    \caption{\textbf{Extended Results of Concept Localization.} (Continued on next page.)}
\end{figure*}

\begin{figure*}[!ht]
    \ContinuedFloat 
    \centering
    \begin{subfigure}{\textwidth}
        \includegraphics[width=\linewidth]{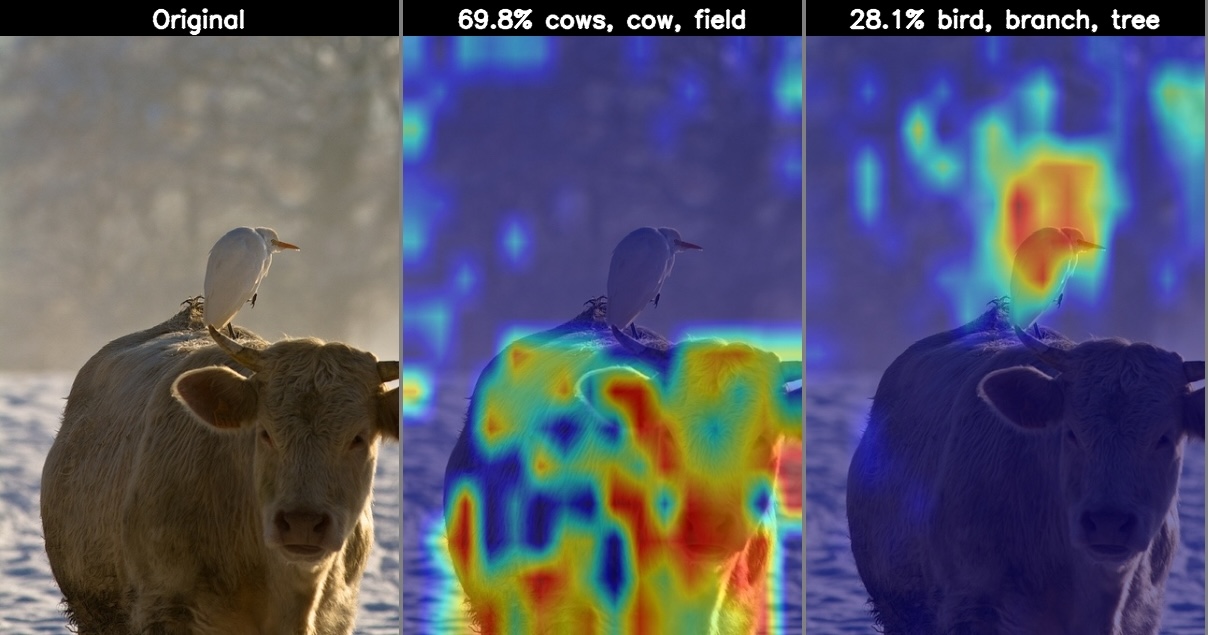}
    \end{subfigure}\vspace{1.5mm}
    \begin{subfigure}{\textwidth}
        \includegraphics[width=\linewidth]{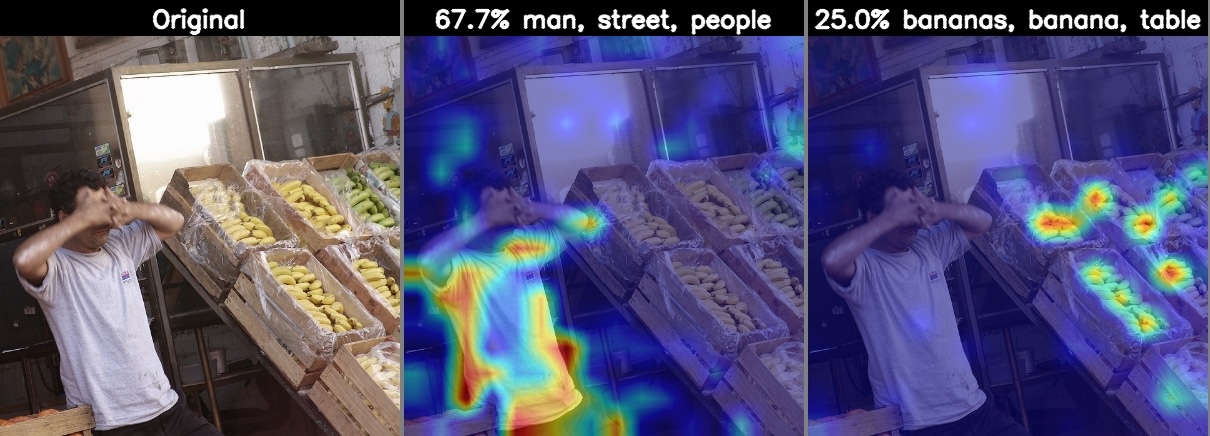}
    \end{subfigure}
    \caption{\textbf{Extended Results of Concept Localization (Continued).} MovMF-CLIP successfully disentangles complex visual scenes into orthogonal semantic components. Each heatmap accurately grounds specific keywords derived from our mixture components, demonstrating precise spatial isolation of objects.}
    \label{fig:supp_heatmaps}
\end{figure*}

\subsection{Semantic Stability under Iterative Generative Drift}
\label{subsec:supp_generative_drift}

We further analyze semantic stability under the iterative generative drift across 6 consecutive steps. We begin by evaluating the quantitative degradation of the semantic representations, followed by a detailed visual analysis.

\begin{figure}[t]
    \centering
    \begin{subfigure}{0.48\linewidth}
        \includegraphics[width=\linewidth]{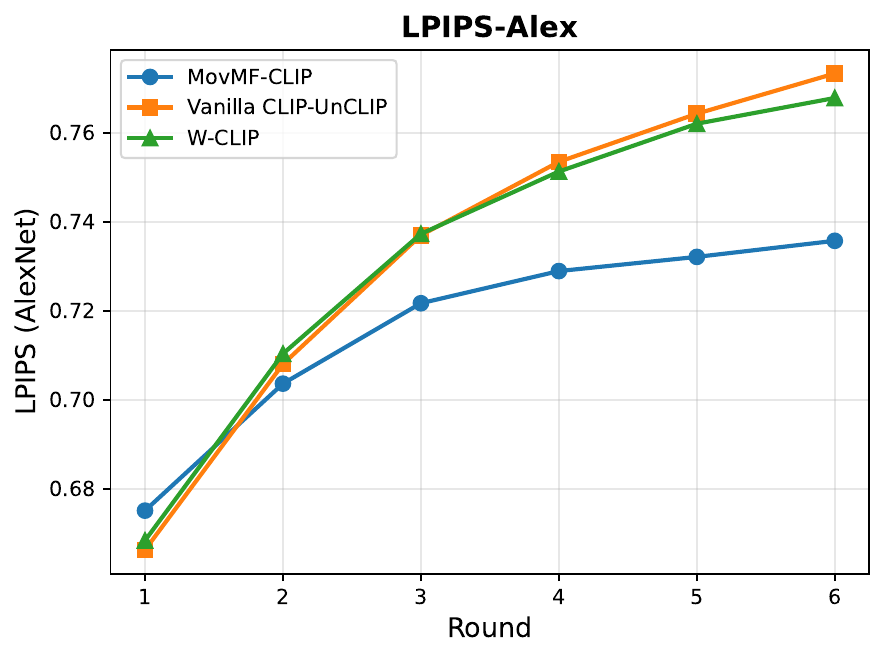} 
    \end{subfigure}
    \hfill
    \begin{subfigure}{0.48\linewidth}
        \includegraphics[width=\linewidth]{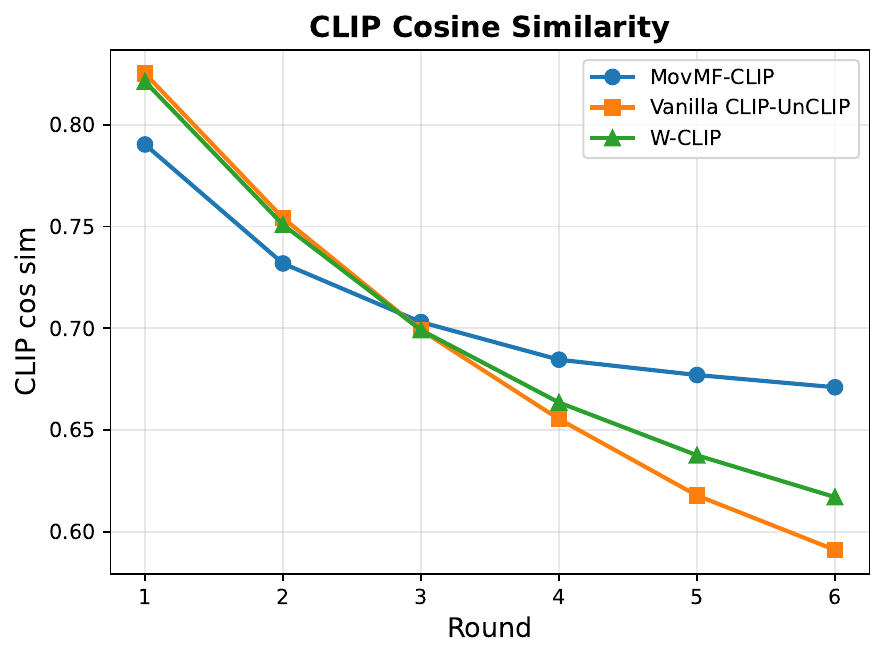} 
    \end{subfigure}
    \caption{\textbf{Quantitative Trends across Iterations.} We track the average LPIPS (left, $\downarrow$) and CLIP Cosine Similarity (right, $\uparrow$) over 1,000 COCO images across 6 CLIP-UnCLIP iterations. While W-CLIP and Vanilla CLIP-UnCLIP show a slight initial advantage, they suffer from accelerating degradation. In contrast, MovMF-CLIP stabilizes the trajectory, significantly outperforming the baseline from round 3 onward.}
    \label{fig:unclip_trends}
\end{figure}

As shown in Fig.~\ref{fig:unclip_trends}, we track the quantitative degradation across 1,000 COCO images. Vanilla CLIP-UnCLIP and W-CLIP initially show slightly lower LPIPS and higher cosine similarity in the first round. However, their representations rapidly deteriorate in subsequent iterations, reflected by a sharp, accelerating drop in CLIP Cosine Similarity and a continuous spike in LPIPS distance. In contrast, our MovMF-CLIP effectively stabilizes the semantic trajectory. By dynamically pulling the latent embeddings toward high-density semantic prototypes, MovMF-CLIP significantly outperforms the baselines from round 3 onward, proving its strong capability in stabilizing representations.

\begin{figure*}[!ht]
    \centering
    \begin{subfigure}{\textwidth}
        \includegraphics[width=\textwidth]{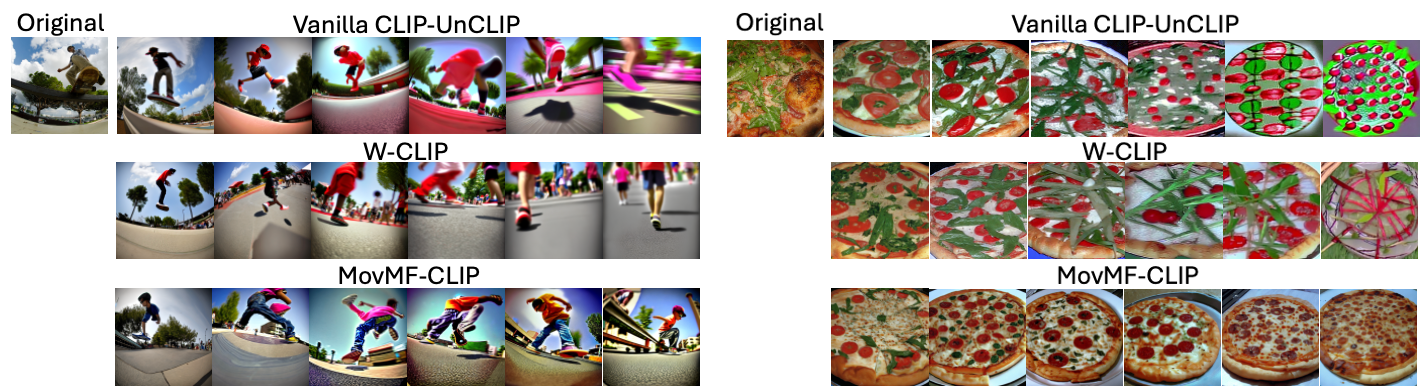} 
    \end{subfigure}
    
    \vspace{3mm}
    
    \begin{subfigure}{\textwidth}
        \includegraphics[width=\textwidth]{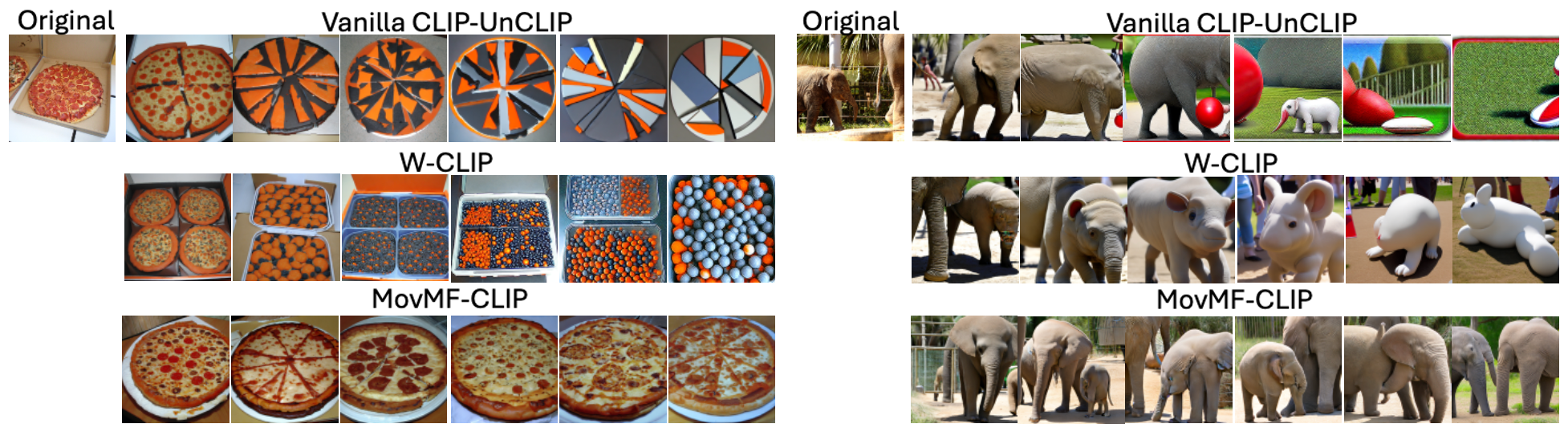} 
    \end{subfigure}
    
    \vspace{3mm}
    
    \begin{subfigure}{\textwidth}
        \includegraphics[width=\textwidth]{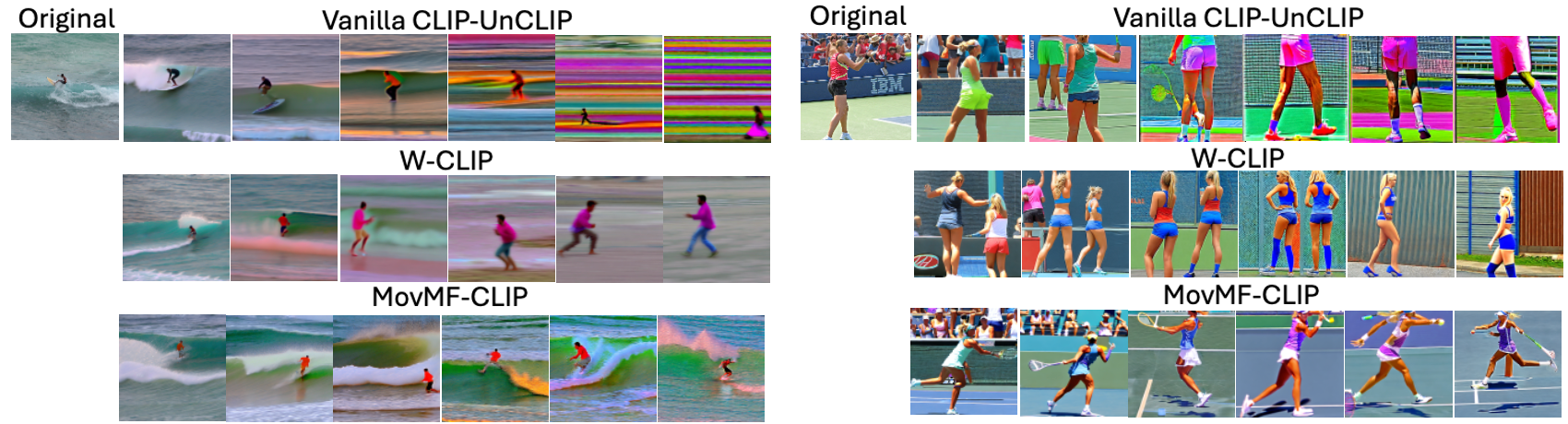} 
    \end{subfigure}
    \caption{\textbf{Extended Results of Iterative Generative Drift.} Trajectories over 6 CLIP-UnCLIP iterations. Vanilla CLIP-UnCLIP suffers from severe mode collapse (\emph{e.g.,} generating meaningless stripes). W-CLIP loses semantic details. Our MovMF-CLIP correctly anchors the embeddings to high-density semantic prototypes, better preserving the original concepts.}
    \label{fig:supp_generative_drift}
\end{figure*}

Fig.~\ref{fig:supp_generative_drift} displays the generated results across 6 consecutive steps. This experiment dramatically exposes the geometric flaws of standard Gaussian assumptions and highlights the power of our MovMF-CLIP manifold projection. For Vanilla CLIP-UnCLIP, representations quickly fall into low-density semantic voids, causing catastrophic mode collapse (\emph{e.g.,} a surfer turning into colorful horizontal stripes, or a pizza collapsing into abstract red and green patterns). While W-CLIP mitigates some noise, it suffers from severe semantic blurring over time, turning elephants into generic smooth blobs and a pizza box into strange blue balls. Conversely, MovMF-CLIP effectively leverages the learned semantic prototypes as geometric anchors. It forcefully locks the latent trajectory, preserving the core semantic identities (skater, pizza, surfer, tennis player, elephant) with remarkable fidelity across all iterations.

\end{document}